\documentclass[english]{article}
\usepackage[T1]{fontenc}
\usepackage[latin9]{inputenc}
\usepackage{geometry}
\usepackage{babel}
\usepackage{verbatim}
\usepackage{mathtools}
\usepackage{amsmath}
\usepackage{amssymb}
\usepackage[unicode=true,
 bookmarks=false,
 breaklinks=false,pdfborder={0 0 1},colorlinks=false]
 {hyperref}
\hypersetup{
 colorlinks,linkcolor=red,anchorcolor=blue,citecolor=blue}

\makeatletter
\usepackage{babel}
\usepackage{babel}
\usepackage{babel}

\usepackage{mathtools}

\usepackage{cite}\usepackage{amsthm}\usepackage{dsfont}\usepackage{array}\usepackage{mathrsfs}\usepackage{comment}\onecolumn
\usepackage{natbib}
\usepackage{color}\usepackage{babel}

\newcommand{\mymid}{\,|\,} 

\allowdisplaybreaks

\usepackage{enumitem}
\setlist[itemize]{leftmargin=1.5em}
\setlist[enumerate]{leftmargin=1.5em}

\usepackage{babel}
\usepackage{algorithm}% http://ctan.org/pkg/algorithms
\usepackage{algorithmic}% http://ctan.org/pkg/algorithmicx
\usepackage{arydshln}

\DeclareMathOperator{\ind}{\mathds{1}}  % Indicator

\numberwithin{equation}{section}

\definecolor{yxc}{RGB}{255,0,0}
\definecolor{yjc}{RGB}{125,0,0}
\definecolor{cm}{RGB}{0,0,200}
\definecolor{yly}{RGB}{0,150,0}

\makeatother

\begin{document}
\theoremstyle{plain} \newtheorem{lemma}{\textbf{Lemma}} \newtheorem{prop}{\textbf{Proposition}}\newtheorem{theorem}{\textbf{Theorem}}\setcounter{theorem}{0}
\newtheorem{corollary}{\textbf{Corollary}} \newtheorem{assumption}{\textbf{Assumption}}
\newtheorem{example}{\textbf{Example}} \newtheorem{definition}{\textbf{Definition}}
\newtheorem{fact}{\textbf{Fact}} \newtheorem{condition}{\textbf{Condition}}\theoremstyle{definition}

\theoremstyle{remark}\newtheorem{remark}{\textbf{Remark}}\newtheorem{claim}{\textbf{Claim}}\newtheorem{conjecture}{\textbf{Conjecture}}
\title{A Score-Based Density Formula, with Applications in \\ Diffusion
Generative Models}
\author{Gen Li\footnote{The authors contributed equally.}
\thanks{Department of Statistics, The Chinese University of Hong Kong, Hong
Kong; Email: \texttt{genli@cuhk.edu.hk}.}\and Yuling Yan\footnotemark[1]
\thanks{Department of Statistics, University of Wisconsin-Madison, Madison, WI 53706, USA; Email: \texttt{yuling.yan@wisc.edu}.}}

\maketitle
\begin{abstract}
Score-based generative models (SGMs) have revolutionized the field
of generative modeling, achieving unprecedented success in generating
realistic and diverse content. Despite empirical advances, the theoretical basis for why optimizing the evidence lower bound (ELBO) on the log-likelihood is effective for training diffusion generative models, such as DDPMs, remains largely unexplored.
In this paper, we address this question by establishing a density formula for a continuous-time diffusion process, which can be viewed as the continuous-time limit of the forward process in an SGM. This formula
reveals the connection between the target density and the score function
associated with each step of the forward process. Building on this, we demonstrate that the minimizer of the optimization objective for training DDPMs nearly coincides with that of the true objective, providing a theoretical foundation for optimizing DDPMs using the ELBO.
Furthermore, we offer new insights into the role of score-matching
regularization in training GANs, the use of ELBO in diffusion classifiers,
and the recently proposed diffusion loss. 
%\yly{Recent research has even integrated the ELBO of pre-trained DDPMs into other generative frameworks, achieving success in GAN training, classification, etc. However, the theoretical justification for this remains unclear.} 
\end{abstract}


\setcounter{tocdepth}{2}

\tableofcontents{}

\section{Introduction}

Score-based generative models (SGMs) represent a groundbreaking advancement
in the realm of generative models, significantly impacting machine
learning and artificial intelligence by their ability to synthesize
high-fidelity data instances, including images, audio, and text \citep{sohl2015deep,ho2020denoising,song2020score,song2019generative,dhariwal2021diffusion,song2020denoising}.
These models operate by progressively refining noisy data into samples
that resemble the target distribution. Due to their innovative approach,
SGMs have achieved unprecedented success, setting new standards in
generative AI and demonstrating extraordinary proficiency in generating
realistic and diverse content across various domains, from image synthesis
and super-resolution to audio generation and molecular design \citep{ramesh2022hierarchical,rombach2022high,saharia2022photorealistic,croitoru2023diffusion,yang2022diffusion}. 

The foundation of SGMs is rooted in the principles of stochastic processes,
especially stochastic differential equations (SDEs). These models
utilize a forward process, which involves the gradual corruption of
an initial data sample with Gaussian noise over several time steps.
This forward process can be described as:
\begin{equation}
X_{0}\overset{\textsf{add noise}}{\longrightarrow}X_{1}\overset{\textsf{add noise}}{\longrightarrow}\cdots\overset{\textsf{add noise}}{\longrightarrow}X_{T},\label{eq:forward-process}
\end{equation}
where $X_{0}\sim p_{\mathsf{data}}$ is the original data sample,
and $X_{T}$ is a sample close to pure Gaussian noise. The ingenuity
of SGMs lies in constructing a reverse denoising process that iteratively
removes the noise, thereby reconstructing the data distribution. This
reverse process starts from a Gaussian sample $Y_{T}$ and moves backward
as:

\begin{equation}
Y_{T}\overset{\textsf{denoise}}{\longrightarrow}Y_{T-1}\overset{\textsf{denoise}}{\longrightarrow}\cdots\overset{\textsf{denoise}}{\longrightarrow}Y_{0}\label{eq:reverse-process}
\end{equation}
ensuring that $Y_{t}\overset{\mathrm{d}}{\approx}X_{t}$ at each step
$t$. The final output $Y_{0}$ is a new sample that closely mimics
the distribution of the initial data $p_{\mathsf{data}}$. 

Inspired by the classical results on time-reversal of SDEs \citep{anderson1982reverse,haussmann1986time},
SGMs construct the reverse process guided by score functions $\nabla\log p_{X_{t}}$
associated with each step of the forward process. Although these score
functions are unknown, they are approximated by neural networks trained
through score-matching techniques \citep{hyvarinen2005estimation,hyvarinen2007some,vincent2011connection,song2019generative}.
This leads to two popular models: denoising diffusion probabilistic
models (DDPMs) \citep{ho2020denoising,nichol2021improved} and denoising
diffusion implicit models (DDIMs) \citep{song2020denoising}. While
the theoretical results in this paper do not depend on the specific
construction of the reverse process, we will use the DDPM framework
to discuss their implications for diffusion generative models.

However, despite empirical advances, there remains a lack of theoretical
understanding for diffusion generative models. For instance, the optimization
target of DDPM is derived from a variational lower bound on
the log-likelihood \citep{ho2020denoising}, which is also
referred to as the evidence lower bound (ELBO) \citep{luo2022understanding}.
It is not yet clear, from a theoretical standpoint, why optimizing
a lower bound of the true objective is still a valid approach. More
surprisingly, recent research suggests incorporating the ELBO
of a pre-trained DDPM into other generative or learning frameworks
to leverage the strengths of multiple architectures, effectively using
it as a proxy for the negative log-likelihood of the data distribution.
This approach has shown empirical success in areas such as GAN training,
classification, and inverse problems \citep{xia2023smart,li2023your,graikos2022diffusion,mardani2024a}.
While it is conceivable that the ELBO is a reasonable optimization
target for training DDPMs (as similar idea is utilized in e.g., the
majorize-minimization algorithm), it is more mysterious why it serves
as a good proxy for the negative log-likelihood in these applications.

In this paper, we take a step towards addressing the aforementioned
question. On the theoretical side, we establish a density formula
for a diffusion process $(X_{t})_{0\leq t<1}$ defined by the following
SDE:
\[
\mathrm{d}X_{t}=-\frac{1}{2(1-t)}X_{t}\mathrm{d}t+\frac{1}{\sqrt{1-t}}\mathrm{d}B_{t}\quad(0\leq t<1),\qquad X_{0}\sim p_{\mathsf{data}},
\]
which can be viewed as a continuous-time limit of the forward process
(\ref{eq:forward-process}). Under some regularity conditions, this formula
expresses the density of $X_{0}$ with the score function along this
process, having the form
\[
\log p_{X_{0}}(x)=-\frac{1+\log(2\pi)}{2}d-\int_{0}^{1}\left[\frac{1}{2(1-t)}\mathbb{E}\Big[\Big\|\frac{X_{t}-\sqrt{1-t}X_{0}}{t}+\nabla\log p_{X_{t}}(X_{t})\Big\|_{2}^{2}\mymid X_{0}=x\Big]-\frac{d}{2t}\right]\mathrm{d}t,
\]
where $p_{X_{t}}(\cdot)$ is the density of $X_{t}$. By time-discretization,
this reveals the connection between the target density $p_{\mathsf{data}}$
and the score function associated with each step of the forward process
(\ref{eq:forward-process}). These theoretical results will be presented
in Section~\ref{sec:main_results}.

Finally, using this density formula, we demonstrate that the minimizer of the optimization target for training DDPMs (derived from the ELBO) also nearly minimizes the true target---the KL divergence between the target distribution and the generator distribution. This finding provides a theoretical foundation for optimizing DDPMs using the ELBO. Additionally, we use this
formula to offer new insights into the role of score-matching regularization
in training GANs \citep{xia2023smart}, the use of ELBO in diffusion
classifiers \citep{li2023your}, and the recently proposed diffusion
loss \citep{li2024autoregressive}. These implications will be discussed
in Section~\ref{sec:implications}.

\section{Problem set-up}

In this section, we formally introduce the Denoising Diffusion Probabilistic
Model (DDPM) and the stochastic differential equation (SDE) that describes
the continuous-time limit of the forward process of DDPM. 

\subsection{Denoising diffusion probabilistic model \label{subsec:ddpm}}

Consider the following forward Markov process in discrete time:
\begin{equation}
X_{t}=\sqrt{1-\beta_{t}}X_{t-1}+\sqrt{\beta_{t}}W_{t}\quad(t=1,\ldots,T),\qquad X_{0}\sim p_{\mathsf{data}},\label{eq:forward-update}
\end{equation}
where $W_{1},\ldots,W_{T}\overset{\text{i.i.d.}}{\sim}\mathcal{N}(0,I_{d})$
and the learning rates $\beta_{t}\in(0,1)$. Since our main results
do not depend on the specific choice of $\beta_{t}$, we will specify
them as needed in later discussions. For each $t\in[T]$, let $q_{t}$
be the law or density function of $X_{t}$, and let $\alpha_{t}\coloneqq1-\beta_{t}$
and $\overline{\alpha}_{t}\coloneqq\prod_{i=1}^{t}\alpha_{i}$. A
simple calculation shows that:
\begin{equation}
X_{t}=\sqrt{\overline{\alpha}_{t}}X_{0}+\sqrt{1-\overline{\alpha}_{t}}\,\overline{W}_{t}\qquad\text{where}\qquad\overline{W}_{t}\sim\mathcal{N}(0,I_{d}).\label{eq:forward-formula}
\end{equation}
We will choose the learning rates $\beta_{t}$ to ensure that $\overline{\alpha}_{T}$
is sufficiently small, such that $q_{T}$ is close to the standard
Gaussian distribution. 

The key components for constructing the reverse process in the context
of DDPM are the score functions $s_{t}^{\star}:\mathbb{R}^{d}\to\mathbb{R}^{d}$
associated with each $q_{t}$, defined as the gradient of their log
density:
\[
s_{t}^{\star}(x)\coloneqq\nabla\log q_{t}(x)\quad(t=1,\ldots,T).
\]
While these score functions are not explicitly known, in practice,
noise-prediction networks $\varepsilon_{t}(x)$ are trained to predict
\[
\varepsilon_{t}^{\star}(x)\coloneqq-\sqrt{1-\overline{\alpha}_{t}}s_{t}^{\star}(x),
\]
which are often referred to as epsilon predictors. To construct the
reverse process, we use:

\begin{equation}
Y_{t-1}=\frac{1}{\sqrt{\alpha_{t}}}\big(Y_{t}+\eta_{t}s_{t}\left(Y_{t}\right)+\sigma_{t}Z_{t}\big)\quad(t=T,\ldots,1),\qquad Y_{T}\sim\mathcal{N}(0,I_{d})\label{eq:DDPM}
\end{equation}
where $Z_{1},\ldots,Z_{T}\overset{\text{i.i.d.}}{\sim}\mathcal{N}(0,I_{d})$,
and $s_{t}(\cdot)\coloneqq-\varepsilon_{t}(\cdot)/\sqrt{1-\overline{\alpha}_{t}}$
is the estimate of the score function $s_{t}^{\star}(\cdot)$. Here
$\eta_{t},\sigma_{t}>0$ are the coefficients that influence the performance
of the DDPM sampler, and we will specify them as needed in later discussion.
For each $t\in[T]$, we use $p_{t}$ to denote the law or density
of $Y_{t}$. 

\subsection{A continuous-time SDE for the forward process \label{subsec:sde}}

In this paper, we build our theoretical results on the continuous-time
limit of the aforementioned forward process, described by the diffusion
process:
\begin{equation}
\mathrm{d}X_{t}=-\frac{1}{2(1-t)}X_{t}\mathrm{d}t+\frac{1}{\sqrt{1-t}}\mathrm{d}B_{t}\quad(0\leq t<1),\qquad X_{0}\sim p_{\mathsf{data}},\label{eq:diffusion-SDE}
\end{equation}
where $(B_{t})_{t\geq0}$ is a standard Brownian motion. The solution
to this stochastic differential equation (SDE) has the closed-form
expression:
\begin{equation}
X_{t}=\sqrt{1-t}X_{0}+\sqrt{t}\,\overline{Z}_{t}\qquad\text{where}\qquad\overline{Z}_{t}=\sqrt{\frac{1-t}{t}}\int_{0}^{t}\frac{1}{1-s}\mathrm{d}B_{s}\sim\mathcal{N}(0,I_{d}).\label{eq:SDE-formula}
\end{equation}
It is important to note that the process $X_{t}$ is not defined at
$t=1$, although it is straightforward to see from the above equation
that $X_{t}$ converges to a Gaussian variable as $t\to1$. 

To demonstrate the connection between this diffusion process and the
forward process (\ref{eq:forward-update}) of the diffusion model,
we evaluate the diffusion process at times $t_{i}=\sqrt{1-\overline{\alpha}_{i}}$
for $1\leq i\leq T$. It is straightforward to check that the marginal
distribution of the resulting discrete-time process $\{X_{t_{i}}:1\leq i\leq T\}$
is identical to that of the forward process (\ref{eq:forward-update}).
Therefore the diffusion process (\ref{eq:diffusion-SDE}) can be viewed
as a continuous-time limit of the forward process. In the next section,
we will establish theoretical results for the diffusion process (\ref{eq:diffusion-SDE}).
Through time discretization, our theory will provide insights for
the DDPM.

We use the notation $X_{t}$ for both the discrete-time process $\{X_{t}:t\in[T]\}$
in (\ref{eq:forward-update}) and the continuous-time diffusion process
$(X_{t})_{0\leq t<1}$ in (\ref{eq:diffusion-SDE}) to maintain consistency
with standard literature. The context will clarify which process is
being referred to.

\section{The score-based density formula\label{sec:main_results}}

\subsection{Main results}

Our main results are based on the continuous-time diffusion process
$(X_{t})_{0\leq t<1}$ defined in (\ref{eq:diffusion-SDE}). While
$X_{0}$ might not have a density, for any $t\in(0,1)$, the random
variable $X_{t}$ has a smooth density, denoted by $\rho_{t}(\cdot)$.
Our main result characterizes the evolution of the conditional mean
of $\log\rho_{t}(X_{t})$ given $X_{0}$, as stated below.

\begin{theorem} \label{thm:main} Consider the diffusion process
$(X_{t})_{0\leq t<1}$ defined in (\ref{eq:diffusion-SDE}), and let
$\rho_{t}$ be the density of $X_{t}$. For any $0<t_{1}<t_{2}<1$,
we have 
\[
\mathbb{E}\left[\log\rho_{t_{2}}(X_{t_{2}})-\log\rho_{t_{1}}(X_{t_{1}})\mymid X_{0}\right]=\int_{t_{1}}^{t_{2}}\bigg(\frac{1}{2(1-t)}\mathbb{E}\Big[\Big\|\frac{X_{t}-\sqrt{1-t}X_{0}}{t}+\nabla\log\rho_{t}(X_{t})\Big\|_{2}^{2}\mymid X_{0}\Big]-\frac{d}{2t}\bigg)\mathrm{d}t.
\]
\end{theorem}

The proof of this theorem is deferred to Appendix~\ref{sec:proof-theorem-main}.
A few remarks are as follows. First, it is worth mentioning that this
formula does not describe the evolution of the (conditional) differential
entropy of the process, because $\rho_{t}(\cdot)$ represents the
unconditional density of $X_{t}$, while the expectation is taken
conditional on $X_{0}$. Second, without further assumptions, we cannot
set $t_{1}=0$ or $t_{2}=1$ because $X_{0}$ might not have a density
(hence $\rho_{0}$ is not well-defined), and $X_{t}$ is only defined
for $t<1$. By assuming that $X_{0}$ has a finite second moment,
the following proposition characterizes the limit of $\mathbb{E}[\log\rho_{t}(X_{t})\mymid X_{0}]$
as $t\to1$.

\begin{prop}\label{prop:limit-1}Suppose that $\mathbb{E}[\Vert X_{0}\Vert_{2}^{2}]<\infty$.
Then for any $x_{0}\in\mathbb{R}^{d}$, we have 
\[
\lim_{t\to1-}\mathbb{E}\left[\log\rho_{t}\left(X_{t}\right)\mymid X_{0}=x_{0}\right]=-\frac{1+\log\left(2\pi\right)}{2}d.
\]
\end{prop}

The proof of this proposition is deferred to Appendix~\ref{sec:proof-prop-limit-1}.
This result is not surprising, as it can be seen from (\ref{eq:SDE-formula})
that $X_{t}$ converges to a standard Gaussian variable as $t\to1$
regardless of $x_{0}$, and we can check
\[
\mathbb{E}[\log\phi(Z)]=-\frac{1+\log\left(2\pi\right)}{2}d
\]
 where $Z\sim\mathcal{N}(0,I_{d})$ and $\phi(\cdot)$ is its density
(we will use this notation throughout his section). The proof of Proposition~\ref{prop:limit-1}
formalizes this intuitive analysis. 

When $X_{0}$ has a smooth density $\rho_{0}(\cdot)$ with Lipschitz
continuous score function, we can show that $\mathbb{E}[\log\rho_{t}(X_{t})\mymid X_{0}]\to\rho_{0}(x_{0})$
as $t\to0$, as presented in the next proposition.

\begin{prop} \label{prop:limit-0} Suppose that $X_{0}$ has density
$\rho_{0}(\cdot)$ and $\sup_{x}\Vert\nabla^{2}\log\rho_{0}(x)\Vert<\infty$.
Then for any $x_{0}\in\mathbb{R}^{d}$, we have 
\[
\lim_{t\to0+}\mathbb{E}\left[\log\rho_{t}\left(X_{t}\right)\mymid X_{0}=x_{0}\right]=\log\rho_{0}(x_{0}).
\]
\end{prop}

The proof of this proposition can be found in Appendix~\ref{sec:proof-prop-limit-0}.
With Propositions~\ref{prop:limit-1}~and~\ref{prop:limit-0} in
place, we can take $t_{1}\to0$ and $t_{2}\to1$ in Theorem~\ref{thm:main}
to show that for any given point $x_{0}$, \begin{subequations}\label{eq:density-formula-all}
\begin{align}
\log\rho_{0}(x_{0}) & =-\frac{1+\log(2\pi)}{2}d-\int_{0}^{1}D(t,x_{0})\mathrm{d}t\label{eq:density-formula-smooth}
\end{align}
where the function $D(x,t)$ is defined as
\begin{equation}
D(t,x)\coloneqq\frac{1}{2(1-t)}\mathbb{E}\Big[\Big\|\frac{X_{t}-\sqrt{1-t}X_{0}}{t}+\nabla\log\rho_{t}(X_{t})\Big\|_{2}^{2}\mymid X_{0}=x\Big]-\frac{d}{2t}.\label{eq:density-formula-integrand}
\end{equation}
In practice, we might not want to make smoothness assumptions on $X_{0}$
as in Proposition~\ref{prop:limit-0}. In that case, we can fix some
sufficiently small $\delta>0$ and obtain a density formula
\begin{align}
\mathbb{E}\left[\log\rho_{\delta}\left(X_{\delta}\right)\mymid X_{0}=x_{0}\right] & =-\frac{1+\log(2\pi)}{2}d-\int_{\delta}^{1}D(t,x_{0})\mathrm{d}t\label{eq:density-formula-general}
\end{align}
\end{subequations}for a smoothed approximation of $\log\rho_{0}(x_{0})$.
This kind of proximity is often used to circumvent non-smoothness
target distributions in diffusion model literature (e.g.,~\citet{li2023towards,chen2022sampling,chen2023probability,benton2023linear}).
We leave some more discussions to Appendix~\ref{appendix:discussion}.

\subsection{From continuous time to discrete time}

In this section, to avoid ambiguity, we will use $(X_{t}^{\mathsf{sde}})_{0\leq t<1}$
to denote the continuous-time diffusion process (\ref{eq:diffusion-SDE})
studied in the previous section, while keep using $\{X_{t}:1\leq t\leq T\}$
to denote the forward process (\ref{eq:forward-update}). The density
formula (\ref{eq:density-formula-all}) is not readily implementable
because of its continuous-time nature. Consider time discretization
over the grid 
\[
0<t_{1}<t_{2}<\cdots<t_{T}<t_{T+1}=1\qquad\text{where}\qquad t_{i}\coloneqq1-\overline{\alpha}_{i}\quad(1\leq i\leq T).
\]
Recall that the forward process $X_{1},\ldots,X_{T}$ has the same
marginal distribution as $X_{t_{1}}^{\mathsf{sde}},\ldots,X_{t_{T}}^{\mathsf{sde}}$
snapshoted from the diffusion process (\ref{eq:diffusion-SDE}). This
gives the following approximation of the density formula (\ref{eq:density-formula-smooth}):
\begin{align*}
\log\rho_{0}(x_{0}) & \overset{\text{(i)}}{\approx}\mathbb{E}\left[\log\rho_{t_{1}}(X_{t_{1}}^{\mathsf{sde}})\mymid X_{0}^{\mathsf{sde}}=x_{0}\right]\\
 & \overset{\text{(ii)}}{\approx}-\frac{1+\log(2\pi t_{1})}{2}d-\sum_{i=1}^{T}\frac{t_{i+1}-t_{i}}{2(1-t_{i})}\mathbb{E}\Big[\Big\|\frac{X_{t_{i}}^{\mathsf{sde}}-\sqrt{1-t_{i}}X_{0}^{\mathsf{sde}}}{t_{i}}+\nabla\log\rho_{t_{i}}(X_{t}^{\mathsf{sde}})\Big\|_{2}^{2}\mymid X_{0}^{\mathsf{sde}}=x_{0}\Big]\\
 & \overset{\text{(iii)}}{\approx}-\frac{1+\log\left(2\pi t_{1}\right)}{2}d-\sum_{i=1}^{T}\frac{t_{i+1}-t_{i}}{2t_{i}(1-t_{i})}\mathbb{E}_{\varepsilon\sim\mathcal{N}(0,I_{d})}\Big[\big\Vert\varepsilon-\widehat{\varepsilon}_{i}(\sqrt{1-t_{i}}x_{0}+\sqrt{t_{i}}\varepsilon)\big\Vert_{2}^{2}\Big].
\end{align*}
In step (i) we approximate $\log\rho_{0}(x_{0})$ with a smoothed
proxy; see the discussion around (\ref{eq:density-formula-general})
for details; step~(ii) applies (\ref{eq:density-formula-general}),
where we compute the integral $\int_{t_{1}}^{1}d/(2t)\mathrm{d}t=-(d/2)\log t_{1}$
in closed form and approximate the integral
\[
\int_{t_{1}}^{1}\frac{1}{2(1-t)}\mathbb{E}\Big[\Big\|\frac{X_{t}^{\mathsf{sde}}-\sqrt{1-t}X_{0}^{\mathsf{sde}}}{t}+\nabla\log\rho_{t}(X_{t}^{\mathsf{sde}})\Big\|_{2}^{2}\mymid X_{0}^{\mathsf{sde}}=x_{0}\Big]\mathrm{d}t;
\]
step (iii) follows from $X_{t_{i}}^{\mathsf{sde}}\overset{\text{d}}{=}\sqrt{1-t_{i}}x_{0}+\sqrt{t_{i}}\varepsilon$
for $\varepsilon\sim\mathcal{N}(0,I_{d})$ conditional on $X_{0}^{\mathsf{sde}}=x_{0}$,
and the relation
\[
\nabla\log\rho_{t_{i}}=\nabla\log q_{i}=s_{i}^{\star}(x)=-\sqrt{t_{i}}\varepsilon_{i}^{\star}(x)\approx-\sqrt{t_{i}}\widehat{\varepsilon}_{i}(x).
\]

In practice, we need to choose the learning rates $\{\beta_{t}:1\leq t\leq T\}$
such that the grid becomes finer as $T$ becomes large. More specifically,
we require 
\[
t_{i+1}-t_{i}=\overline{\alpha}_{i}-\overline{\alpha}_{i+1}=\overline{\alpha_{i}}\beta_{i+1}\leq\beta_{i+1}\quad(1\leq i\leq T-1)
\]
to be small (roughly of order $O(1/T)$), and $t_{1}=\beta_{1}$ and
$1-t_{T}=\overline{\alpha}_{T}$ to be vanishingly small (of order
$T^{-c}$ for some sufficiently large constant $c>0$); see e.g.,
\citet{li2023towards,benton2023linear} for learning rate schedules
satisfying these properties. Finally, we replace the time steps $\{t_{i}:1\leq i\leq T\}$
with the learning rates for the forward process to achieve\footnote{Here we define $\alpha_{T+1}=0$ to accommodate the last term in the summation.}
\begin{align}
\log\rho_{0}(x_{0}) & \approx-\frac{1+\log\left(2\pi\beta_{1}\right)}{2}d-\sum_{t=1}^{T}\frac{1-\alpha_{t+1}}{2(1-\overline{\alpha}_{t})}\mathbb{E}_{\varepsilon\sim\mathcal{N}(0,I_{d})}\Big[\big\Vert\varepsilon-\widehat{\varepsilon}_{t}(\sqrt{\overline{\alpha}_{t}}x_{0}+\sqrt{1-\overline{\alpha}_{t}}\varepsilon)\big\Vert_{2}^{2}\Big],\label{eq:density-formula-discrete}
\end{align}
The density approximation (\ref{eq:density-formula-discrete}) can
be evaluated with the trained epsilon predictors. 

\subsection{Comparison with other results}

The density formulas (\ref{eq:density-formula-all}) expresses the
density of $X_{0}$ using the score function along the continuous-time
limit of the forward process of the diffusion model. Other forms of
score-based density formulas can be derived using normalizing flows.
Notice that the probability flow ODE of the SDE (\ref{eq:diffusion-SDE})
is
\begin{equation}
\dot{x}_{t}=v_{t}(x_{t})\qquad\text{where}\qquad v_{t}(x)=-\frac{x-\nabla\log\rho_{t}(x)}{2(1-t)};\label{eq:prob-ode}
\end{equation}
namely, if we draw a particle $x_{0}\sim\rho_{0}$ and evolve it according
to the ODE (\ref{eq:prob-ode}) to get the trajectory $t\to x_{t}$
for $t\in[0,1)$, then $x_{t}\sim\rho_{t}$. See e.g., \citet[Appendix D.1]{song2020score}
for the derivation of this result. 

Under some smoothness condition, we can use the results developed
in \citet{grathwohl2018scalable,albergo2023stochastic} to show that
for any given $x_{0}$
\begin{equation}
\log\rho_{t}(x_{t})-\log\rho_{0}(x_{0})=-\int_{0}^{t}\mathsf{Tr}\left(\frac{\partial}{\partial x}v_{s}(x_{s})\right)\mathrm{d}s=\int_{0}^{t}\frac{d-\mathsf{tr}\big(\nabla^{2}\log\rho_{s}(x_{s})\big)}{2(1-s)}\mathrm{d}s.\label{eq:density-formula-grathwohl-1}
\end{equation}
Here $t\to x_{t}$ is the solution to the ODE (\ref{eq:prob-ode})
with initial condition $x_{0}$. Since the ODE system (\ref{eq:prob-ode})
is based on the score functions (hence $x_{t}$ can be numerically
solved), and the integral in (\ref{eq:density-formula-grathwohl-1})
is based on the Jacobian of the score functions, we may take $t\to1$
and use the fact that $\rho_{t}(\cdot)\to\phi(\cdot)$ to obtain a
score-based density formula
\begin{equation}
\log\rho_{0}(x_{0})=-\frac{d}{2}\log(2\pi)-\frac{1}{2}\left\Vert x_{1}\right\Vert _{2}^{2}-\int_{0}^{1}\frac{d-\mathsf{tr}\big(\nabla^{2}\log\rho_{s}(x_{s})\big)}{2(1-s)}\mathrm{d}s.\label{eq:density-formula-grathwohl-2}
\end{equation}
However, numerically, this formula is more difficult to compute than
our formula (\ref{eq:density-formula-all}) for the following reasons.
First, (\ref{eq:density-formula-grathwohl-2}) involves the Jacobian
of the score functions, which are more challenging to estimate than
the score functions themselves. In fact, existing convergence guarantees
for DDPM do not depend on the accurate estimation of the Jacobian
of the score functions \citep{benton2023linear,chen2022improved,chen2022sampling,li2024adapting}.
Second, using this density formula requires solving the ODE (\ref{eq:prob-ode})
accurately to obtain $x_{1}$, which might not be numerically stable,
especially when the score function is not accurately estimated at
early stages, due to error propagation. In contrast, computing (\ref{eq:density-formula-all})
only requires evaluating a few Gaussian integrals (which can be efficiently
approximated by the Monte Carlo method) and is more stable to score
estimation error. 

\section{Implications \label{sec:implications}}

In the previous section, we established a density formula 
\begin{align}
\log q_{0}(x) & \approx\underbrace{-\frac{1+\log\left(2\pi\beta_{1}\right)}{2}d}_{\eqqcolon C_{0}^{\star}}-\sum_{t=1}^{T}\underbrace{\frac{1-\alpha_{t+1}}{2(1-\overline{\alpha}_{t})}\mathbb{E}_{\varepsilon\sim\mathcal{N}(0,I_{d})}\Big[\big\Vert\varepsilon-\varepsilon_{t}^{\star}(\sqrt{\overline{\alpha}_{t}}x+\sqrt{1-\overline{\alpha}_{t}}\varepsilon)\big\Vert_{2}^{2}\Big]}_{\eqqcolon L_{t-1}^{\star}(x)}\label{eq:our-approx}
\end{align}
up to discretization error (which vanishes as $T$ becomes large)
and score estimation error. In this section, we will discuss the implications
of this formula in various generative and learning frameworks.

\subsection{Certifying the validity of optimizing ELBO in DDPM \label{subsec:vlb}}

The seminal work \citep{ho2020denoising} established the variational
lower bound (VLB), also known as the evidence lower bound (ELBO),
of the log-likelihood
\begin{align}
\log p_{0}(x) & \geq-\sum_{t=2}^{T}\underbrace{\mathbb{E}_{x_{t}\sim p_{X_{t}|X_{0}}(\cdot\mymid x)}\mathsf{KL}\left(p_{X_{t-1}|X_{t},X_{0}}(\cdot\mymid x_{t},x)\,\Vert\,p_{Y_{t-1}|Y_{t}}(\cdot\mymid x_{t})\right)}_{\eqqcolon L_{t-1}(x)}\nonumber \\
 & \qquad-\underbrace{\mathsf{KL}\left(p_{Y_{T}}(\cdot)\,\Vert\,p_{X_{T}|X_{0}}(\cdot\mymid x)\right)}_{\eqqcolon L_{T}(x)}+\underbrace{\mathbb{E}_{x_{1}\sim p_{X_{1}|X_{0}}(\cdot\mymid x)}\left[\log p_{Y_{0}|Y_{1}}(x\mymid x_{1})\right]}_{\eqqcolon C_{0}(x)},\label{eq:vlb}
\end{align}
where the reverse process $(Y_{t})_{0\leq t\leq T}$ was defined in
Section~\ref{subsec:ddpm}, and $p_{0}$ is the density of $Y_{0}$.
Under the coefficient design recommended by \citet{li2024adapting}
(other reasonable designs also lead to similar conclusions) 
\begin{equation}
\eta_{t}=1-\alpha_{t}\qquad\text{and}\qquad\sigma_{t}^{2}=\frac{\left(1-\alpha_{t}\right)\left(\alpha_{t}-\overline{\alpha}_{t}\right)}{1-\overline{\alpha}_{t}},\label{eq:defn-step-size}
\end{equation}
it can be computed that for each $2\leq t\leq T$: 
\[
L_{t-1}(x)=\frac{1-\alpha_{t}}{2(\alpha_{t}-\overline{\alpha}_{t})}\mathbb{E}_{\varepsilon\sim\mathcal{N}(0,I_{d})}\left[\big\Vert\varepsilon-\varepsilon_{t}(\sqrt{\overline{\alpha}_{t}}x+\sqrt{1-\overline{\alpha}_{t}}\varepsilon)\big\Vert_{2}^{2}\right].
\]
We can verify that (i) for each $2\leq t\leq T$, the coefficients
in $L_{t-1}$ from (\ref{eq:vlb}) and $L_{t-1}^{\star}$ from (\ref{eq:our-approx})
are identical up to higher-order error; (ii) when $T$ is large, $L_{T}$
becomes vanishingly small; and (iii) the function 
\[
C_{0}(x)=-\frac{1+\log\left(2\pi\beta_{1}\right)}{2}d+O(\beta_{1})=C_{0}^{\star}+O(\beta_{1})
\]
is nearly a constant. See Appendix~\ref{appendix:vlb} for details.
It is worth highlighting that as far as we know, existing literature
haven't pointed out that $C_{0}(x)$ is nearly a constant. For instance,
\citet{ho2020denoising} discretize this term to obtain discrete log-likelihood
(see Section 3.3 therein), which is unnecessary in view of our observation.
Additionally, some later works falsely claim that $C_{0}(x)$ is negligible,
as we will discuss in the following sections.

Now we discuss the validity of optimizing the variational bound for training DDPMs.
Our discussion shows that 
\begin{align}
\underbrace{\mathsf{KL}(q_{0}\parallel p_{0})}_{\eqqcolon\mathcal{L}(\varepsilon_{1},\ldots,\varepsilon_{T})} & =-\mathbb{E}_{x\sim q_{0}}[\log p_{0}(x)]-H(q_{0})\leq\underbrace{\mathbb{E}_{x\sim q_{0}}[L(x)]-C_{0}^{\star}-H(q_{0})+o(1)}_{\eqqcolon\mathcal{L}_{\mathsf{vb}}(\varepsilon_{1},\ldots,\varepsilon_{T})},\label{eq:vlb-all}
\end{align}
where $H(q_{0})=-\int\log q_{0}(x)\mathrm{d}q_{0}$ is the entropy
of $q_{0}$, and $L(x)$ denotes the widely used (negative) ELBO\footnote{We follow the convention in existing literature to remove the last
two terms $L_{T}(x)$ and $C_{0}(x)$ from \eqref{eq:vlb} in the
ELBO.} 
\[
L(x)\coloneqq\sum_{t=1}^{T}\frac{1-\alpha_{t+1}}{2(1-\overline{\alpha}_{t})}\mathbb{E}_{\varepsilon\sim\mathcal{N}(0,I_{d})}\Big[\big\Vert\varepsilon-\varepsilon_{t}(\sqrt{\overline{\alpha}_{t}}x+\sqrt{1-\overline{\alpha}_{t}}\varepsilon)\big\Vert_{2}^{2}\Big].
\]
The true objective of DDPM is to learn the epsilon predictors $\varepsilon_{1},\ldots,\varepsilon_{T}$
that minimizes $\mathcal{L}$ in (\ref{eq:vlb-all}),
while in practice, the optimization target is the variational bound $\mathcal{L}_{\mathsf{vb}}$.
It is known that the global minimizer for 
\begin{equation}
\mathbb{E}_{x\sim q_{0}}\left[L(x)\right]=\sum_{t=1}^{T}\frac{1-\alpha_{t+1}}{2(1-\overline{\alpha}_{t})}\mathbb{E}_{x\sim q_{0},\varepsilon\sim\mathcal{N}(0,I_{d})}\Big[\big\Vert\varepsilon-\varepsilon_{t}(\sqrt{\overline{\alpha}_{t}}x+\sqrt{1-\overline{\alpha}_{t}}\varepsilon)\big\Vert_{2}^{2}\Big]\label{eq:expected-elbo}
\end{equation}
is exactly $\widehat{\varepsilon}_{t}(\cdot)\equiv\varepsilon_{t}^{\star}(\cdot)$
for each $1\leq t\leq T$ (see Appendix~\ref{appendix:vlb}). Although
in practice the optimization is based on samples from the target
distribution $q_{0}$ (instead of the population level expectation
over $q_{0}$) and may not find the exact global minimizer, we consider the ideal scenario where the learned epsilon predictors $\widehat{\varepsilon}_{t}$
equal $\varepsilon_{t}^{\star}$ to facilitate discussion. When
$\varepsilon_{t}=\varepsilon_{t}^{\star}$ for each $t$, according
to (\ref{eq:our-approx}), we have 
\begin{equation}
L(x)\approx-\log q_{0}(x)+C_{0}^{\star}.\label{eq:elbo}
\end{equation}
Taking \eqref{eq:vlb-all} and \eqref{eq:elbo} together gives
\begin{equation}
0 \leq \mathcal{L} (\widehat{\varepsilon}_{1},\ldots,\widehat{\varepsilon}_{T}) \leq \mathcal{L}_{\mathsf{vb}}(\widehat{\varepsilon}_{1},\ldots,\widehat{\varepsilon}_{T})\approx-\mathbb{E}_{x\sim q_{0}}[\log q(x)]+C_{0}^{\star}-C_{0}^{\star}-H(q_{0}) = 0,\label{eq:elbo-approx-1}
\end{equation}
namely the minimizer for $\mathcal{L}_{\mathsf{vb}}$ approximately minimizes $\mathcal{L}$, and the optimal value is asymptotically zero when the number of steps $T$ becomes large. This suggests that by minimizing the variational bound $\mathcal{L}_{\mathsf{vb}}$, the resulting generator distribution $p_0$ is guaranteed to be close to the target distribution $q_0$ in KL divergence.

Some experimental evidence suggests that using reweighted coefficients
can marginally improve empirical performance. For example, \citet{ho2020denoising}
suggests that in practice, it might be better to use uniform coefficients
in the ELBO 
\begin{align}
L_{\mathsf{simple}}(x) & \coloneqq\frac{1}{T}\sum_{i=1}^{T}\mathbb{E}_{\varepsilon\sim\mathcal{N}(0,I_{d})}\Big[\big\Vert\varepsilon-\widehat{\varepsilon}_{t_{i}}(\sqrt{\overline{\alpha}_{t}}x+\sqrt{1-\overline{\alpha}_{t}}\varepsilon)\big\Vert_{2}^{2}\Big]\label{eq:elbo-simplified}
\end{align}
when trainging DDPM to improve sampling quality.\footnote{Note that the optimal epsilon predictors $\widehat{\varepsilon}_{t}$
for $L$ and $L_{\mathsf{simple}}$ are the same, but in practice,
we may not find the optimal predictors. This practical strategy is
beyond the scope of our theoretical result, and implies that the influence
of terms from different steps needs more careful investigation. We
conjecture that this is mainly because the estimation error for terms
when $t$ is close to zero is larger, hence smaller coefficients for
these terms can improve performance.} This strategy has been adopted by many later works. In the following
sections, we will discuss the role of using the ELBO in different
applications. While the original literature might use the modified
ELBO (\ref{eq:elbo-simplified}), in our discussion we will stick
to the original ELBO (\ref{eq:elbo}) to gain intuition from our theoretical
findings.

\subsection{Understanding the role of regularization in GAN \label{subsec:gan}}

Generative Adversarial Networks (GANs) are a powerful and flexible
framework for learning the unknown probability distribution $p_{\mathsf{data}}$
that generates a collection of training data \citep{goodfellow2014generative}.
GANs operate on a game between a generator $G$ and a discriminator
$D$, typically implemented using neural networks. The generator $G$
takes a random noise vector $z$ sampled from a simple distribution
$p_{\mathsf{noise}}$ (e.g., Gaussian) and maps it to a data sample
resembling the training data, aiming for the distribution of $G(z)$
to be close to $p_{\mathsf{data}}$. Meanwhile, the discriminator
$D$ determines whether a sample $x$ is real (i.e., drawn from $p_{\mathsf{data}}$)
or fake (i.e., produced by the generator), outputting the probability
$D(x)$ of the former. The two networks engage in a zero-sum game:
\[
\min_{G}\max_{D}V(G,D)\coloneqq\mathbb{E}_{x\sim p_{\mathsf{data}}}[\log D(x)]+\mathbb{E}_{z\sim p_{\mathsf{noise}}}[\log(1-D(G(z)))],
\]
with the generator striving to produce realistic data while the discriminator
tries to distinguish real data from fake. The generator and discriminator
are trained iteratively\footnote{While the most natural update rule for the generator is $G\leftarrow\arg\min\,\mathbb{E}_{z\sim p_{\mathsf{noise}}}[\log(1-D(G(z)))]$,
both schemes are used in practice and have similar performance. Our
choice is for consistency with \citet{xia2023smart}, and our analysis
can be extended to the other choice.} 
\begin{align*}
D & \leftarrow\arg\min\,-\mathbb{E}_{x\sim p_{\mathsf{data}}}[\log D(x)]-\mathbb{E}_{z\sim p_{\mathsf{noise}}}[\log(1-D(G(z)))],\\
G & \leftarrow\arg\min\,-\mathbb{E}_{z\sim p_{\mathsf{noise}}}[\log D(G(z))]
\end{align*}
to approach the Nash equilibrium $(G^{\star},D^{\star})$, where the
distribution of $G^{\star}(z)$ with $z\sim p_{\mathsf{noise}}$ matches
the target distribution $p_{\mathsf{data}}$, and $D(x)=1/2$ for
all $x$.

It is believed that adding a regularization term to make the generated
samples fit the VLB can improve the sampling quality of the generative
model. For example, \citet{xia2023smart} proposed adding the VLB
$L(x)$ as a regularization term to the objective function, where
$\{\widehat{\varepsilon}_{t_{i}}(\cdot):1\leq i\leq T\}$ are the
learned epsilon predictors for $p_{\mathsf{data}}$. The training
procedure then becomes 
\begin{align*}
D & \leftarrow\arg\min\,-\mathbb{E}_{x\sim p_{\mathsf{data}}}[\log D(x)]-\mathbb{E}_{z\sim p_{\mathsf{noise}}}[\log(1-D(G(z)))],\\
G & \leftarrow\arg\min\,-\mathbb{E}_{z\sim p_{\mathsf{noise}}}[\log D(G(z))]+\lambda\mathbb{E}_{z\sim p_{\mathsf{noise}}}[L(G(z))],
\end{align*}
where $\lambda>0$ is some tuning parameter. However, it remains unclear
what exactly is optimized through the above objective. According to
our theory, $L(x)\approx-\log p_{\mathsf{data}}(x)+C_{0}^{\star}.$
Assuming that this approximation is exact for intuitive understanding,
the unique Nash equilibrium $(G_{\lambda},D_{\lambda})$ satisfies
\begin{align*}
p_{G_{\lambda}}(x) & =\big(zp_{\mathsf{data}}(x)^{\lambda}-1\big)_{+}p_{\mathsf{data}}(x)
\end{align*}
for some normalizing factor $z>0$, where $p_{G_{\lambda}}$ is the
density of $G_{\lambda}(z)$ with $z\sim p_{\mathsf{noise}}$. See
Appendix~\ref{appendix:gan} for details. This can be viewed as amplifying
the density $p_{\mathsf{data}}$ wherever it is not too small, while
zeroing out the density where $p_{\mathsf{data}}$ is vanishingly
small (which is difficult to estimated accurately), thus improving
the sampling quality.

\subsection{Confirming the use of ELBO in diffusion classifier}

Motivated by applications like image classification and text-to-image
diffusion model, we consider a joint underlying distribution $p_{0}(x,c)$,
where typically $x$ is the image data and the latent variable $c$
is the class index or text embedding, taking values in a finite set
$\mathcal{C}$. For each $c\in\mathcal{C}$, we train a diffusion
model for the conditional data distribution $p_{0}(x\mymid c)$, which
provides a set of epsilon predictors $\big\{\widehat{\varepsilon}_{t}(x;c):1\leq t\leq T,c\in\mathcal{C}\big\}$.
Assuming a uniform prior over $\mathcal{C}$, we can use Bayes' formula
to obtain: 
\[
p_{0}\left(c\mymid x\right)=\frac{p_{0}\left(c\right)p_{0}\left(x\mymid c_{i}\right)}{\sum_{j\in\mathcal{C}}p_{0}\left(c_{j}\right)p_{0}\left(x\mymid c_{j}\right)}=\frac{p_{0}\left(x\mymid c\right)}{\sum_{j\in\mathcal{C}}p_{0}\left(x\mymid c_{j}\right)}.
\]
for each $c\in\mathcal{C}$. Recent work \citep{li2023your} proposed
to use the ELBO\footnote{The original paper adopted uniform coefficients; see the last paragraph
of Section~\ref{subsec:vlb} for discussion.} 
\[
-L(x;c)\coloneqq-\sum_{t=1}^{T}\frac{1-\alpha_{t+1}}{2(1-\overline{\alpha}_{t})}\mathbb{E}_{\varepsilon\sim\mathcal{N}(0,I_{d})}\Big[\big\Vert\varepsilon-\widehat{\varepsilon}_{t}(\sqrt{\overline{\alpha}_{t}}x+\sqrt{1-\overline{\alpha}_{t}}\varepsilon;c)\big\Vert_{2}^{2}\Big]
\]
as an approximate class-conditional log-likelihood $\log p_{0}(x\mymid c)$
for each $c\in\mathcal{C}$, which allows them to obtain a posterior
distribution 
\begin{equation}
\widehat{p}_{0}\left(c\mymid x\right)=\frac{\exp\left(-L(x;c)\right)}{\sum_{j\in\mathcal{C}}\exp\left(-L(x;c_{j})\right)}.\label{eq:posterior}
\end{equation}
Our theory suggests that $-L(x;c)\approx\log p_{0}(x\mymid c)-C_{0}^{\star}$,
where $C_{0}^{\star}=-[1+\log(2\pi\beta_{1})]d/2$ is a universal
constant that does not depend on $p_{0}$ and $c$. This implies that
\[
\widehat{p}_{0}\left(c\mymid x\right)\approx\frac{\exp\left(\log p_{0}(x\mymid c)-C_{0}^{\star}\right)}{\sum_{j\in\mathcal{C}}\exp\left(\log p_{0}(x\mymid c_{j})-C_{0}^{\star}\right)}=\frac{p_{0}\left(x\mymid c\right)}{\sum_{j\in\mathcal{C}}p_{0}\left(x\mymid c_{j}\right)}=p_{0}\left(c\mymid x\right)
\]
providing theoretical justification for using the computed posterior
$\widehat{p}_{0}$ in classification tasks.

It is worth mentioning that, although this framework was proposed
in the literature \citep{li2023your}, it remains a heuristic method
before our work. For example, in general, replacing the intractable
log-likelihood with a lower bound does not guarantee good performance,
as they might not be close. Additionally, recall that there is a term
$C_{0}(x)$ in the ELBO (\ref{eq:vlb}). \citet{li2023your} claimed
that ``\emph{Since $T=1000$ is large and $\log p_{\theta}(x_{0}\mymid x_{1},c)$
is typically small, we choose to drop this term}''. However this
argument is not correct, as we already computed in Section~\ref{subsec:vlb}
that this term 
\[
C_{0}(x)=-\frac{1+\log\left(2\pi\beta_{1}\right)}{2}d+O(\beta_{1})
\]
can be very large since $\beta_{1}$ is typically very close to $0$.
In view of our results, the reason why this term can be dropped is
that it equals a universal constant that does not depend on the image
data $x$ and the class index $c$, thus it does not affect the posterior
(\ref{eq:posterior}).

\subsection{Demystifying the diffusion loss in autoregressive models}

Finally, we use our results to study a class of diffusion loss recently
introduced in \citet{li2024autoregressive}, in the context of autoregressive
image generation. Let $x^{k}$ denote the next token to be predicted,
and $z$ be the condition parameterized by an autoregressive network
$z=f(x^{1},\ldots,x^{k-1})$ based on previous tokens as input. %For each $z$, suppose that we have already trained a diffusion model
%for $p(x^{k}\mymid z)$, i.e., the probability distribution of $x^{k}$
%conditioned on $z$, which returns a set of epsilon predictors $\{\widehat{\varepsilon}_{t}(\cdot\,;z):1\leq t\leq T\}$.
%The goal is to train the network $f(\cdot)$ such that $p(x\mymid z)$
%with $z=f(x^{1},\ldots,x^{k-1})$ can predict the next token $x^{k}$.
The goal is to train the network $z=f(\cdot)$ together with a diffusion
model $\{\varepsilon_{t}(\cdot\,;z):1\leq t\leq T\}$ such
that $\widehat{p}(x\mymid z)$ (induced by the diffusion model)
with $z=f(x^{1},\ldots,x^{k-1})$ can predict the next token $x^{k}$.

The diffusion loss is defined as follows: for some weights $w_{t}\ge0$,
let 
\begin{align}
L(z,x) & =\sum_{t=1}^{T}w_{t}\mathbb{E}_{\varepsilon\sim\mathcal{N}(0,I_{d})}\Big[\big\Vert\varepsilon-\varepsilon_{t}(\sqrt{\overline{\alpha}_{t}}x+\sqrt{1-\overline{\alpha}_{t}}\varepsilon;z)\big\Vert_{2}^{2}\Big].\label{eq:diffusion_loss-1}
\end{align}
With training data $\{(x_{i}^{1},\ldots,x_{i}^{k}):1\leq i\leq n\}$,
we can train the autoregressive network $f(\cdot)$ and the diffusion model by minimizing
the following empirical risk: 
\begin{equation}
\mathop{\arg\min}_{f,\varepsilon_1,\ldots,\varepsilon_T}\frac{1}{n}\sum_{i=1}^{n}L\left(f(x_{i}^{1},\ldots,x_{i}^{k-1}),x_{i}^{k}\right).\label{eq:diffusion_loss-2}
\end{equation}
To gain intuition from our theoretical results, we take the weights
in the diffusion loss (\ref{eq:diffusion_loss-1}) to be the coefficients
in the ELBO (\ref{eq:elbo}), and for each $z$, suppose that the
learned diffusion model for $p(x^{k}\mymid z)$ is already good enough, which
returns the set of epsilon predictors $\{\widehat{\varepsilon}_{t}(\cdot\,;z):1\leq t\leq T\}$
for the probability distribution of $x^{k}$ conditioned on $z$.
Under this special case, our approximation result (\ref{eq:elbo})
shows that 
\[
L(z,x)\approx-\log p(x\mymid z)+C_{0}^{\star},
\]
which suggests that the training objective for the network $f$ in (\ref{eq:diffusion_loss-2})
can be viewed as approximate MLE, as the loss function 
\[
\frac{1}{n}\sum_{i=1}^{n}L\left(f(x_{i}^{1},\ldots,x_{i}^{k-1}),x_{i}^{k}\right)\approx-\frac{1}{n}\sum_{i=1}^{n}\log p(x_{i}^{k}\mymid f(x_{i}^{1},\ldots,x_{i}^{k-1}))+C_{0}^{\star}
\]
represents the negative log-likelihood function (up to an additive
constant) of the observed $x_{1}^{k},\ldots,x_{n}^{k}$ in terms of
$f$.

\begin{comment}
In \cite{li2024autoregressive}, the authors claim that this diffusion
loss ``\textit{conceptually behaves like a form of score matching:
it is related to a loss function concerning the score function of
$p(x|z)$}''. To gain intuition from our theoretical results, we
take the weights in the diffusion loss (\ref{eq:diffusion_loss-1})
to be the coefficients in the ELBO (\ref{eq:elbo}). Notice that the
diffusion loss (\ref{eq:diffusion_loss-1}) is defined for one sample
$x$. When applied to samples generated from a distribution $q$,
the expected diffusion loss 
\[
\mathbb{E}_{x\sim q}[L(z,x)]\approx-\int q(x)\log p(x\mymid z)\mathrm{d}x+C_{0}^{\star}=\mathsf{KL}\big(q\,\Vert\,p(\cdot\mymid z)\big)+H(q)+C_{0}^{\star},
\]
where the first relation follows from our approximation result (\ref{eq:elbo}),
and $H(q)\coloneqq-\int q(x)\log q(x)\mathrm{d}x$ denotes the entropy
of $q$. Minimizing the diffusion loss amounts to optimizing the distribution
$q$ and the autoregressive network $z=f(\cdot)$ jointly. The above
formula implies that the diffusion loss can promote the generated
samples (from the distribution $q$) to be closer to the target distribution
$p(\cdot\mymid z)$, but with the cost of decreasing diversity due
to the entropy term $H(q)$. 
\end{comment}

\section{Proof of Theorem~\ref{thm:main} \label{sec:proof-theorem-main}}

Recall the definition of the stochastic process $(X_{t})_{0\leq t\leq1}$
\[
\mathrm{d}X_{t}=-\frac{1}{2(1-t)}X_{t}\mathrm{d}t+\frac{1}{\sqrt{1-t}}\mathrm{d}B_{t}.
\]
Define $Y_{t}\coloneqq X_{t}/\sqrt{1-t}$ for any $0\leq t<1$, and
let $f(t,x)=x/\sqrt{1-t}$, we can use Itô's formula to show that
\begin{align}
\mathrm{d}Y_{t} & =\mathrm{d}f\left(t,X_{t}\right)=\frac{\partial f}{\partial t}\left(t,X_{t}\right)\mathrm{d}t+\nabla_{x}f\left(t,X_{t}\right)^{\top}\mathrm{d}X_{t}+\frac{1}{2}\mathrm{d}X_{t}^{\top}\nabla_{x}^{2}f\left(t,X_{t}\right)\mathrm{d}X_{t}\nonumber \\
 & =\frac{X_{t}}{2(1-t)^{3/2}}\mathrm{d}t+\frac{1}{\sqrt{1-t}}\left(-\frac{1}{2(1-t)}X_{t}\mathrm{d}t+\frac{1}{\sqrt{1-t}}\mathrm{d}B_{t}\right)=\frac{\mathrm{d}B_{t}}{1-t}.\label{eq:dYt-dBt}
\end{align}
Therefore the Itô process $Y_{t}$ is a martingale, which is easier
to handle. Let $g(t,y)=\log\rho_{t}(\sqrt{1-t}y)$, and we can express
$\log\rho_{t}(x)=g(t,x/\sqrt{1-t})$. In view of Itô's formula, we
have 
\begin{align}
\mathrm{d}\log\rho_{t}(X_{t}) & =\mathrm{d}g\left(t,Y_{t}\right)\overset{\text{(i)}}{=}\frac{\partial g}{\partial t}\left(t,Y_{t}\right)\mathrm{d}t+\nabla_{y}g\left(t,Y_{t}\right)^{\top}\mathrm{d}Y_{t}+\frac{1}{2}\mathrm{d}Y_{t}^{\top}\nabla_{y}^{2}g\left(t,Y_{t}\right)\mathrm{d}Y_{t}\nonumber \\
 & \overset{\text{(ii)}}{=}\frac{\partial g}{\partial t}\left(t,Y_{t}\right)\mathrm{d}t+\frac{1}{1-t}\nabla_{y}g\left(t,Y_{t}\right)^{\top}\mathrm{d}B_{t}+\frac{1}{2\left(1-t\right)^{2}}\mathrm{d}B_{t}^{\top}\nabla_{y}^{2}g\left(t,Y_{t}\right)\mathrm{d}B_{t}\nonumber \\
 & \overset{\text{(iii)}}{=}\frac{\partial g}{\partial t}\left(t,Y_{t}\right)\mathrm{d}t+\frac{1}{1-t}\nabla_{y}g\left(t,Y_{t}\right)^{\top}\mathrm{d}B_{t}+\frac{1}{2\left(1-t\right)^{2}}\mathsf{tr}\left(\nabla_{y}^{2}g\left(t,Y_{t}\right)\right)\mathrm{d}t.\label{eq:proof-main-1}
\end{align}
Here step (i) follows from the Itô rule, step (ii) utilizes (\ref{eq:dYt-dBt}),
while step (iii) can be derived from the Itô calculus. Then we investigate
the three terms above. Notice that 
\begin{align}
\nabla_{y}g\left(t,y\right)\mymid_{y=Y_{t}} & =\frac{\nabla_{y}\rho_{t}(\sqrt{1-t}y)}{\rho_{t}(\sqrt{1-t}Y_{t})}\mymid_{y=Y_{t}}=\frac{\nabla_{x}\rho_{t}(X_{t})\sqrt{1-t}}{\rho_{t}(X_{t})}=\sqrt{1-t}\nabla\log\rho_{t}\left(X_{t}\right),\label{eq:proof-main-2}
\end{align}
and similarly, we have 
\begin{equation}
\nabla_{y}^{2}g\left(t,y\right)\mymid_{y=Y_{t}}=\left(1-t\right)\nabla^{2}\log\rho_{t}\left(X_{t}\right).\label{eq:proof-main-3}
\end{equation}
Substituting (\ref{eq:proof-main-2}) and (\ref{eq:proof-main-3})
back into (\ref{eq:proof-main-1}) gives 
\begin{align*}
\mathrm{d}\log\rho_{t}(X_{t}) & =\frac{\partial g}{\partial t}\left(t,Y_{t}\right)\mathrm{d}t+\frac{1}{\sqrt{1-t}}\nabla\log\rho_{t}\left(X_{t}\right)^{\top}\mathrm{d}B_{t}+\frac{1}{2\left(1-t\right)}\mathsf{tr}\left(\nabla^{2}\log\rho_{t}\left(X_{t}\right)\right)\mathrm{d}t.
\end{align*}
or equivalently, for any given $0<t_{1}<t_{2}<1$, we have 
\begin{equation}
\log\rho_{t}\left(X_{t}\right)\Big|_{t_{1}}^{t_{2}}=\int_{t_{1}}^{t_{2}}\Big[\frac{\partial g}{\partial t}\left(t,Y_{t}\right)+\frac{\mathsf{tr}\left(\nabla^{2}\log\rho_{t}\left(X_{t}\right)\right)}{2\left(1-t\right)}\Big]\mathrm{d}t+\int_{t_{1}}^{t_{2}}\frac{1}{\sqrt{1-t}}\nabla\log\rho_{t}\left(X_{t}\right)^{\top}\mathrm{d}B_{t}.\label{eq:proof-main-4}
\end{equation}
Conditional on $X_{0}$, we take expectation on both sides of (\ref{eq:proof-main-4})
to achieve 
\begin{equation}
\mathbb{E}\left[\log\rho_{t_{2}}\left(X_{t_{2}}\right)-\log\rho_{t_{1}}\left(X_{t_{1}}\right)\mymid X_{0}\right]=\mathbb{E}\left[\int_{t_{1}}^{t_{2}}\left(\frac{\partial g}{\partial t}\left(t,Y_{t}\right)+\frac{1}{2\left(1-t\right)}\mathsf{tr}\left(\nabla^{2}\log\rho_{t}\left(X_{t}\right)\right)\right)\mathrm{d}t\mymid X_{0}\right].\label{eq:proof-main-5}
\end{equation}
We need the following lemmas, whose proof can be found at the end
of this section.

\begin{claim}\label{claim-1}For any $0<t<1$ and any $y\in\mathbb{R}^{d}$,
we have 
\[
\frac{\partial g}{\partial t}\left(t,y\right)=-\frac{d}{2t}+\frac{1}{2t^{2}}\int_{x_{0}}\rho_{X_{0}|X_{t}}\left(x_{0}\mymid\sqrt{1-t}y\right)\|y-x_{0}\|_{2}^{2}\mathrm{d}x_{0}.
\]

\end{claim}

\begin{claim}\label{claim-2}For any $0<t<1$ and any $x\in\mathbb{R}^{d}$,
we have 
\begin{align*}
\mathsf{tr}\left(\nabla^{2}\log\rho_{t}(x)\right) & =-\frac{d}{t}-\big\Vert\nabla\log\rho_{t}(x)\big\Vert_{2}^{2}+\frac{1}{t^{2}}\int\big\Vert x-\sqrt{1-t}x_{0}\big\Vert_{2}^{2}\rho_{X_{0}|X_{t}}\left(x_{0}\mymid x\right)\mathrm{d}x_{0}.
\end{align*}
It also admits the lower bound 
\[
\mathsf{tr}\left(\nabla^{2}\log\rho_{t}(x)\right)\geq-\frac{d}{t}.
\]

\end{claim}

Therefore for any $x$ and $y=x/\sqrt{1-t}$, we know that 
\begin{align}
\frac{\partial g}{\partial t}\left(t,y\right)+\frac{1}{2\left(1-t\right)}\mathsf{tr}\left(\nabla^{2}\log\rho_{t}\left(x\right)\right) & \geq-\frac{d}{2t}-\frac{d}{2\left(1-t\right)t}\geq-\frac{d}{\left(1-t\right)t}.\label{eq:proof-main-6}
\end{align}
Hence we have 
\begin{align}
 & \mathbb{E}\left[\log\rho_{t_{2}}\left(X_{t_{2}}\right)-\log\rho_{t_{1}}\left(X_{t_{1}}\right)\mymid X_{0}\right]\nonumber \\
 & \qquad\overset{\text{(i)}}{=}\mathbb{E}\left[\int_{t_{1}}^{t_{2}}\left(\frac{\partial g}{\partial t}\left(t,Y_{t}\right)+\frac{1}{2\left(1-t\right)}\mathsf{tr}\left(\nabla^{2}\log\rho_{t}\left(X_{t}\right)\right)+\frac{d}{\left(1-t\right)t}\right)\mathrm{d}t\mymid X_{0}\right]-\int_{t_{1}}^{t_{2}}\frac{d}{\left(1-t\right)t}\mathrm{d}t\nonumber \\
 & \qquad\overset{\text{(ii)}}{=}\int_{t_{1}}^{t_{2}}\mathbb{E}\left[\left(\frac{\partial g}{\partial t}\left(t,Y_{t}\right)+\frac{1}{2\left(1-t\right)}\mathsf{tr}\left(\nabla^{2}\log\rho_{t}\left(X_{t}\right)\right)+\frac{d}{\left(1-t\right)t}\right)\mymid X_{0}\right]\mathrm{d}t-\int_{t_{1}}^{t_{2}}\frac{d}{\left(1-t\right)t}\mathrm{d}t\nonumber \\
 & \qquad=\int_{t_{1}}^{t_{2}}\mathbb{E}\left[\left(\frac{\partial g}{\partial t}\left(t,Y_{t}\right)+\frac{1}{2\left(1-t\right)}\mathsf{tr}\left(\nabla^{2}\log\rho_{t}\left(X_{t}\right)\right)\right)\mymid X_{0}\right]\mathrm{d}t.\label{eq:proof-main-7}
\end{align}
Here step (i) follows from (\ref{eq:proof-main-5}), and its validity
is guaranteed by 
\[
\int_{t_{1}}^{t_{2}}\frac{d}{t\left(1-t\right)}\mathrm{d}t=\log\frac{t_{2}\left(1-t_{1}\right)}{t_{1}\left(1-t_{2}\right)}<+\infty,
\]
while step (ii) utilizes Tonelli's Theorem, and the nonnegativity
of the integrand is ensured by (\ref{eq:proof-main-6}). Taking Claims~\ref{claim-1}~and~\ref{claim-2}
collectively, we know that for any $x$ and $y=x/\sqrt{1-t}$, 
\begin{align}
\frac{\partial g}{\partial t}\left(t,y\right)-\frac{\mathsf{tr}\left(\nabla^{2}\log\rho_{t}\left(x\right)\right)}{2\left(1-t\right)} & =\frac{d+\big\Vert\nabla\log\rho_{t}(x)\big\Vert_{2}^{2}}{2\left(1-t\right)}+\frac{1}{2t^{2}}\int_{x_{0}}\rho_{X_{0}|X_{t}}\left(x_{0}\mymid\sqrt{1-t}y\right)\|y-x_{0}\|_{2}^{2}\mathrm{d}x_{0}\nonumber \\
 & \mathbb{\qquad}-\frac{1}{2\left(1-t\right)}\frac{1}{t^{2}}\int\big\Vert x-\sqrt{1-t}x_{0}\big\Vert_{2}^{2}\rho_{X_{0}|X_{t}}\left(x_{0}\mymid x\right)\mathrm{d}x_{0}\nonumber \\
 & =\frac{d+\big\Vert\nabla\log\rho_{t}(x)\big\Vert_{2}^{2}}{2\left(1-t\right)}.\label{eq:proof-main-8}
\end{align}
Putting (\ref{eq:proof-main-7}) and (\ref{eq:proof-main-8}) together,
we arrive at 
\begin{equation}
\mathbb{E}\left[\log\rho_{t_{2}}\left(X_{t_{2}}\right)-\log\rho_{t_{1}}\left(X_{t_{1}}\right)\mymid X_{0}\right]=\int_{t_{1}}^{t_{2}}\mathbb{E}\bigg[\frac{d+\big\Vert\nabla\log\rho_{t}(X_{t})\big\Vert_{2}^{2}}{2\left(1-t\right)}+\frac{1}{1-t}\mathsf{tr}\left(\nabla^{2}\log\rho_{t}\left(X_{t}\right)\right)\mymid X_{0}\bigg]\mathrm{d}t.\label{eq:proof-main-9}
\end{equation}
Notice that conditional on $X_{0}$, we have $X_{t}\sim\mathcal{N}(\sqrt{1-t}X_{0},tI_{d})$.
Then we have 
\begin{align*}
 & \mathbb{E}\left[\log\rho_{t_{2}}\left(X_{t_{2}}\right)-\log\rho_{t_{1}}\left(X_{t_{1}}\right)\mymid X_{0}\right]\\
 & \qquad\overset{\text{(i)}}{=}\int_{t_{1}}^{t_{2}}\mathbb{E}\bigg[\frac{d+\big\Vert\nabla\log\rho_{t}(X_{t})\big\Vert_{2}^{2}}{2\left(1-t\right)}+\frac{1}{1-t}\nabla\log\rho_{t}(X_{t})^{\top}\frac{X_{t}-\sqrt{1-t}X_{0}}{t}\mymid X_{0}\bigg]\mathrm{d}t\\
 & \qquad\overset{\text{(ii)}}{=}\int_{t_{1}}^{t_{2}}\bigg(\frac{1}{2(1-t)}\mathbb{E}\Big[\Big\|\frac{X_{t}-\sqrt{1-t}X_{0}}{t}+\nabla\log\rho_{t}(X_{t})\Big\|_{2}^{2}\mymid X_{0}\Big]-\frac{d}{2t}\bigg)\mathrm{d}t
\end{align*}
Here step (i) follows from (\ref{eq:proof-main-9}) and an application
of Stein's lemma 
\begin{align*}
\mathbb{E}\bigg[\nabla\log\rho_{t}(X_{t})^{\top}\left(X_{t}-\sqrt{1-t}X_{0}\right)\mymid X_{0}\bigg] & =t\mathbb{E}\left[\mathsf{tr}\big(\nabla^{2}\log\rho_{t}(X_{t})\big)\mymid X_{0}\right],
\end{align*}
while step (ii) holds since 
\[
\mathbb{E}\Big[\Big\|\frac{X_{t}-\sqrt{1-t}X_{0}}{t}\Big\Vert_{2}^{2}\Big]=\frac{d}{t}.
\]

\paragraph{Proof of Claim~\ref{claim-1}.}

For any $t\in(0,1)$, since $X_{t}=\sqrt{1-t}X_{0}+\sqrt{t}Z$, we
have 
\begin{equation}
\rho_{t}(\sqrt{1-t}y)=\int_{x_{0}}(2\pi t)^{-d/2}\exp\Big(-\frac{(1-t)\|y-x_{0}\|_{2}^{2}}{2t}\Big)\rho_{0}(\mathrm{d}x_{0}).\label{eq:pt-rescaled-density}
\end{equation}
Note that here $\rho_{0}(\cdot)$ stands for the law of $X_{0}$.
Hence we have 
\begin{align*}
\frac{\partial g}{\partial t}\left(t,y\right) & =\frac{\partial}{\partial t}\log\rho_{t}(\sqrt{1-t}y)=\text{\ensuremath{\frac{1}{\rho_{t}(\sqrt{1-t}y)}}}\frac{\partial}{\partial t}\rho_{t}(\sqrt{1-t}y)\\
 & =\text{\ensuremath{\frac{1}{\rho_{t}(\sqrt{1-t}y)}}}\int_{x_{0}}(2\pi)^{-d/2}\bigg[-\frac{d}{2}t^{-d/2-1}\exp\Big(-\frac{(1-t)\|y-x_{0}\|_{2}^{2}}{2t}\Big)\\
 & \qquad\qquad\qquad\qquad\qquad\qquad\qquad+t^{-d/2}\exp\Big(-\frac{(1-t)\|y-x_{0}\|_{2}^{2}}{2t}\Big)\frac{\|y-x_{0}\|_{2}^{2}}{2t^{2}}\bigg]\rho_{0}(\mathrm{d}x_{0})\\
 & =\text{\ensuremath{\frac{1}{\rho_{t}(\sqrt{1-t}y)}}}\int_{x_{0}}\rho_{X_{t}|X_{0}}\left(\sqrt{1-t}y\mymid x_{0}\right)\left[-\frac{d}{2t}+\frac{\|y-x_{0}\|_{2}^{2}}{2t^{2}}\right]\rho_{0}(\mathrm{d}x_{0})\\
 & =\int_{x_{0}}\Big(-\frac{d}{2t}+\frac{\|y-x_{0}\|_{2}^{2}}{2t^{2}}\Big)\rho_{X_{0}|X_{t}}\left(\mathrm{d}x_{0}\mymid\sqrt{1-t}y\right)
\end{align*}
as claimed.

\paragraph{Proof of Claim~\ref{claim-2}.}

Notice that we can express 
\begin{align*}
\nabla\log\rho_{t}(x) & =-\frac{1}{t}\mathbb{E}\left[X_{t}-\sqrt{1-t}X_{0}\mymid X_{t}=x\right]=-\frac{1}{t}\int_{x_{0}}\left(x-\sqrt{1-t}x_{0}\right)\rho_{X_{0}|X_{t}}\left(\mathrm{d}x_{0}\mymid x\right);
\end{align*}
see \citet{chen2022sampling} for the proof of this relationship. Then
we can compute 
\begin{align*}
\nabla^{2}\log\rho_{t}(x) & =-\frac{1}{t}\Big\{ I_{d}+\frac{1}{t}\mathbb{E}\left[X_{t}-\sqrt{1-t}X_{0}\mymid X_{t}=x\right]\mathbb{E}\left[X_{t}-\sqrt{1-t}X_{0}\mymid X_{t}=x\right]^{\top}\\
 & \qquad\qquad-\frac{1}{t}\mathbb{E}\left[\left(X_{t}-\sqrt{1-t}X_{0}\right)\left(X_{t}-\sqrt{1-t}X_{0}\right)^{\top}\mymid X_{t}=x\right]\Big\}\\
 & =-\frac{1}{t}\Big\{ I_{d}+\frac{1}{t}\Big[\int\left(x-\sqrt{1-t}x_{0}\right)\rho_{X_{0}|X_{t}}\left(\mathrm{d}x_{0}\mymid x\right)\Big]\Big[\int\left(x-\sqrt{1-t}x_{0}\right)\rho_{X_{0}|X_{t}}\left(\mathrm{d}x_{0}\mymid x\right)\Big]^{\top}\\
 & \qquad\qquad-\frac{1}{t}\int\left(x-\sqrt{1-t}x_{0}\right)\left(x-\sqrt{1-t}x_{0}\right)^{\top}\rho_{X_{0}|X_{t}}\left(\mathrm{d}x_{0}\mymid x\right)\Big\}.
\end{align*}
Hence we have 
\begin{align*}
\mathsf{tr}\left(\nabla^{2}\log\rho_{t}(x)\right) & =-\frac{1}{t}\bigg\{ d+\frac{1}{t}\big\Vert\int\left(x-\sqrt{1-t}x_{0}\right)\rho_{X_{0}|X_{t}}\left(\mathrm{d}x_{0}\mymid x\right)\big\Vert_{2}^{2}-\frac{1}{t}\int\big\Vert x-\sqrt{1-t}x_{0}\big\Vert_{2}^{2}\rho_{X_{0}|X_{t}}\left(\mathrm{d}x_{0}\mymid x\right)\bigg\}\\
 & =-\frac{d}{t}-\frac{1}{t^{2}}\big\Vert\nabla\log\rho_{t}(x)\big\Vert_{2}^{2}+\frac{1}{t^{2}}\int\big\Vert x-\sqrt{1-t}x_{0}\big\Vert_{2}^{2}\rho_{X_{0}|X_{t}}\left(x_{0}\mymid x\right)\mathrm{d}x_{0}.
\end{align*}
By Jensen's inequality, we know that 
\[
\mathsf{tr}\left(\nabla^{2}\log\rho_{t}(x)\right)\geq-\frac{d}{t}.
\]

\section{Discussion}

This paper develops a score-based density formula that expresses the
density function of a target distribution using the score function
along a continuous-time diffusion process that bridges this distribution
and standard Gaussian. By connecting this diffusion process with the
forward process of score-based diffusion models, our results provide
theoretical support for training DDPMs by optimizing the ELBO, 
and offer novel insights into several applications of diffusion
models, including GAN training and diffusion classifiers.

Our work opens several directions for future research. 
%For instance, our theoretical results are established for the continuous-time diffusion process. 
First, our theoretical results are established for the continuous-time
diffusion process. It is crucial to carefully analyze the error induced
by time discretization, which could inform the number of steps required
for the results in this paper to be valid in practice. Additionally,
while our results provide theoretical justification for using the
ELBO (\ref{eq:elbo}) as a proxy for the negative log-likelihood of
the target distribution, they do not cover other practical variants
of ELBO with modified weights (e.g., the simplified ELBO (\ref{eq:elbo-simplified})).
Extending our analysis to other diffusion processes might yield new
density formulas incorporating these modified weights. Lastly, further
investigation is needed into other applications of this score-based
density formula, including density estimation and inverse problems.

\section*{Acknowledgements}

G.~Li is supported in part by the Chinese University of Hong Kong
Direct Grant for Research. Y.~Yan was supported in part by a Norbert
Wiener Postdoctoral Fellowship from MIT.

\appendix

\section{Proof of Proposition~\ref{prop:limit-1} \label{sec:proof-prop-limit-1}}

We establish the desired result by sandwiching $\mathbb{E}[\log\rho_{t}(X_{t})\mymid X_{0}=x_{0}]$
and find its limit as $t\to1$ . We first record that the density
of $X_{t}$ can be expressed as 
\begin{equation}
\rho_{t}\left(x\right)=\mathbb{E}_{X_{0}}\bigg[(2\pi t)^{-d/2}\exp\bigg(-\frac{\|x-\sqrt{1-t}X_{0}\|_{2}^{2}}{2t}\bigg)\bigg],\label{eq:pt-density}
\end{equation}
since $X_{t}\overset{\text{d}}{=}\sqrt{1-t}X_{0}+\sqrt{t}Z$ for an
independent variable $Z\sim\mathcal{N}(0,I_{d})$.

\paragraph{Lower bounding $\mathbb{E}[\log\rho_{t}(X_{t})\mymid X_{0}=x_{0}]$. }

Starting from (\ref{eq:pt-density}), for any $x\in\mathbb{R}^{d}$
and any $0<t<1$, 
\begin{align*}
\log\rho_{t}(x) & =\log\mathbb{E}_{X_{0}}\bigg[(2\pi t)^{-d/2}\exp\bigg(-\frac{\|x-\sqrt{1-t}X_{0}\|_{2}^{2}}{2t}\bigg)\bigg]\\
 & \overset{\text{(i)}}{\geq}\log\bigg\{(2\pi t)^{-d/2}\exp\bigg(-\mathbb{E}_{X_{0}}\bigg[\frac{\|x-\sqrt{1-t}X_{0}\|_{2}^{2}}{2t}\bigg]\bigg)\bigg\}\\
 & =-\frac{d}{2}\log(2\pi t)-\mathbb{E}_{X_{0}}\bigg[\frac{\|x-\sqrt{1-t}X_{0}\|_{2}^{2}}{2t}\bigg]\\
 & =-\frac{d}{2}\log(2\pi t)-\frac{\|x\|_{2}^{2}}{2t}-\frac{1-t}{2t}\mathbb{E}[\Vert X_{0}\Vert_{2}^{2}]+\frac{\sqrt{1-t}}{t}\mathbb{E}[x^{\top}X_{0}]\\
 & \overset{\text{(ii)}}{=}-\frac{d}{2}\log(2\pi t)-\left(1+O(\sqrt{1-t})\right)\frac{\|x\|_{2}^{2}}{2t}+O(\sqrt{1-t})\mathbb{E}[\Vert X_{0}\Vert_{2}^{2}].
\end{align*}
Here step (i) follows from Jensen's inequality and the fact that $e^{-x}$
is a convex function, while step (ii) follows from elementary inequalities
\[
\left|\mathbb{E}[x^{\top}X_{0}]\right|\leq\mathbb{E}\big[\Vert x\Vert\Vert X_{0}\Vert_{2}\big]\leq\frac{1}{2}\mathbb{E}\big[\Vert x\Vert_{2}^{2}+\Vert X_{0}\Vert_{2}^{2}\big].
\]
This immediately gives, for any given $x_{0}\in\mathbb{R}^{d}$ and
any $0<t<1$, \begin{subequations}\label{eq:lb} 
\begin{equation}
\mathbb{E}[\log\rho_{t}(X_{t})\mymid X_{0}=x_{0}]\geq\underbrace{-\frac{d}{2}\log(2\pi t)-\frac{1+O(\sqrt{1-t})}{2t}\mathbb{E}\big[\|X_{t}\|_{2}^{2}\mymid X_{0}=x_{0}\big]+O(\sqrt{1-t})\mathbb{E}[\Vert X_{0}\Vert_{2}^{2}]}_{\eqqcolon f_{x_{0}}(t)}.
\end{equation}
Since $\mathbb{E}[\Vert X_{0}\Vert_{2}^{2}]<\infty$, it is straightforward
to check that 
\begin{align}
\lim_{t\to1-}f_{x_{0}}(t) & =-\frac{d}{2}\log(2\pi)-\lim_{t\to1-}\frac{1}{2}\mathbb{E}\left[\Vert\sqrt{1-t}x_{0}+\sqrt{t}Z\Vert_{2}^{2}\right]\qquad\text{for }Z\sim\mathcal{N}(0,I_{d})\nonumber \\
 & =-\frac{d}{2}\log(2\pi)-\frac{d}{2}.
\end{align}
\end{subequations}

\paragraph{Upper bounding $\mathbb{E}[\log\rho_{t}(X_{t})\mymid X_{0}=x_{0}]$. }

Towards that, we need to obtain point-wise upper bound for $\log\rho_{t}(x)$.
Since the desired result only depends on the limiting behavior when
$t\to1$, from now on we only consider $t>0.9$, under which 
\[
(1-t)^{1/4}<\frac{1}{2}\sqrt{\log\frac{1}{1-t}}
\]
holds. It would be helpful to develop the upper bound for the following
two cases separately. 
\begin{itemize}
\item For any $(1-t)^{1/4}<\Vert x\Vert_{2}<0.5\sqrt{\log1/(1-t)}$, we
have 
\begin{align}
\log\rho_{t}(x) & \overset{\text{(a)}}{\leq}\log\mathbb{E}_{X_{0}}\bigg[(2\pi t)^{-d/2}\exp\bigg(-\frac{(\|x\|_{2}-(1-t)^{1/4})^{2}}{2t}\bigg)+\ind\big(\|X_{0}\|_{2}>(1-t)^{-1/4}\big)\bigg]\nonumber \\
 & \overset{\text{(b)}}{\leq}-\frac{d}{2}\log(2\pi t)-\frac{(\|x\|_{2}-(1-t)^{1/4})^{2}}{2t}+\exp\Big(\frac{(\|x\|_{2}-(1-t)^{1/4})^{2}}{2t}\Big)\mathbb{P}\big(\|X_{0}\|_{2}>(1-t)^{-1/4}\big)\nonumber \\
 & \overset{\text{(c)}}{\leq}-\frac{d}{2}\log(2\pi t)-\frac{(\|x\|_{2}-(1-t)^{1/4})^{2}}{2t}+\exp\Big(\frac{\|x\|_{2}^{2}}{2t}\Big)\mathbb{E}[\|X_{0}\|_{2}^{2}](1-t)^{1/2}\nonumber \\
 & \overset{\text{(d)}}{\leq}-\frac{d}{2}\log(2\pi t)-\frac{(\|x\|_{2}-(1-t)^{1/4})^{2}}{2t}+\mathbb{E}[\|X_{0}\|_{2}^{2}](1-t)^{1/4}.\label{eq:log-pt-ub-1}
\end{align}
Here step (a) follows from (\ref{eq:pt-density}); step (b) holds
since $\log(x+y)\leq\log x+y/x$ holds for any $x>0$ and $y\geq0$;
step (c) follows from $\Vert x\Vert_{2}>(1-t)^{1/4}$ and Chebyshev's
inequality; while step (d) holds since $\Vert x\Vert_{2}<0.5\sqrt{\log1/(1-t)}$. 
\item For $\Vert x\Vert_{2}\geq0.5\sqrt{\log1/(1-t)}$ or $\Vert x\Vert\leq(1-t)^{1/4}$,
we will use the naive upper bound 
\begin{align}
\log\rho_{t}(x) & \leq-\frac{d}{2}\log(2\pi t)<0,\label{eq:log-pt-ub-2}
\end{align}
where the first relation simply follows from (\ref{eq:pt-density})
and the second relation holds when $t>0.9$. 
\end{itemize}
Then we have 
\begin{align*}
 & \mathbb{E}[\log\rho_{t}(X_{t})\mymid X_{0}=x_{0}]\overset{\text{(i)}}{\leq}\mathbb{E}[\log\rho_{t}(X_{t})\ind\left\{ (1-t)^{1/4}<\Vert X_{t}\Vert_{2}<0.5\sqrt{\log1/(1-t)}\right\} \mymid X_{0}=x_{0}]\\
 & \qquad\overset{\text{(ii)}}{\leq}\mathbb{E}\bigg[\Big(-\frac{d}{2}\log(2\pi t)-\frac{(\|x\|_{2}-(1-t)^{1/4})^{2}}{2t}+\mathbb{E}[\|X_{0}\|_{2}^{2}](1-t)^{1/4}\Big)\\
 & \qquad\qquad\qquad\cdot\ind\Big\{(1-t)^{1/4}<\Vert X_{t}\Vert_{2}<0.5\sqrt{\log1/(1-t)}\Big\}\mymid X_{0}=x_{0}\bigg]\\
 & \qquad=\underbrace{\Big(-\frac{d}{2}\log(2\pi t)+\mathbb{E}[\|X_{0}\|_{2}^{2}](1-t)^{1/4}\Big)\mathbb{P}\left((1-t)^{1/4}<\Vert X_{t}\Vert_{2}<0.5\sqrt{\log1/(1-t)}\right)}_{\eqqcolon\overline{g}_{x_{0}}(t)}\\
 & \qquad\qquad-\underbrace{\mathbb{E}\left[\frac{(\|X_{t}\|_{2}-(1-t)^{1/4})^{2}}{2t}\ind\left\{ (1-t)^{1/4}<\Vert X_{t}\Vert_{2}<0.5\sqrt{\log1/(1-t)}\right\} \mymid X_{0}=x_{0}\right]}_{\eqqcolon\widetilde{g}_{x_{0}}(t)}.
\end{align*}
Here step (i) follows from (\ref{eq:log-pt-ub-2}), while step (ii)
utilizes (\ref{eq:log-pt-ub-1}). Since $X_{t}$ is a continuous random
variable for any $t\in(0,1)$, we have 
\[
\lim_{t\to1-}\mathbb{P}\left((1-t)^{1/4}<\Vert X_{t}\Vert_{2}<0.5\sqrt{\log1/(1-t)}\right)=1.
\]
Therefore we know that 
\[
\lim_{t\to1-}\overline{g}_{x_{0}}(t)=-\frac{d}{2}\log(2\pi).
\]
Recall that $X_{t}\overset{\text{d}}{=}\sqrt{1-t}X_{0}+\sqrt{t}Z$
for a Gaussian variable $Z\sim\mathcal{N}(0,I_{d})$ independent of
$X_{0}$, we can express 
\begin{align*}
\widetilde{g}_{x_{0}}(t) & =\mathbb{E}\left[\frac{(\|\sqrt{t}Z+\sqrt{1-t}x_{0}\|_{2}-(1-t)^{1/4})^{2}}{2t}\ind\left\{ (1-t)^{1/4}<\Vert\sqrt{t}Z+\sqrt{1-t}x_{0}\Vert_{2}<\frac{1}{2}\sqrt{\log1/(1-t)}\right\} \right]\\
 & =\int\underbrace{\frac{(\|\sqrt{t}z+\sqrt{1-t}x_{0}\|_{2}-(1-t)^{1/4})^{2}}{2t}\ind\bigg\{(1-t)^{1/4}<\Vert\sqrt{t}z+\sqrt{1-t}x_{0}\Vert_{2}<\frac{1}{2}\sqrt{\log\frac{1}{1-t}}\bigg\}\phi(z)}_{\eqqcolon h_{t}(z)}\mathrm{d}z,
\end{align*}
where $\phi(z)=(2\pi)^{-d/2}\exp(-\Vert z\Vert_{2}^{2}/2)$ is the
density function of $\mathcal{N}(0,I_{d})$. For any $t\in(0.9,1)$,
we have 
\[
h_{t}(z)\leq\|\sqrt{t}z+\sqrt{1-t}x_{0}\|_{2}^{2}\phi(z)\leq2(\|z\|_{2}^{2}+\|x_{0}\|_{2}^{2})\phi(z)\eqqcolon h(z),
\]
and it is straightforward to check that 
\[
\int h(z)\mathrm{d}z=2d+2\|x_{0}\|_{2}^{2}<\infty.
\]
By dominated convergence theorem, we know that 
\[
\lim_{t\to1-}\widetilde{g}_{x_{0}}(t)=\lim_{t\to1-}\int h_{t}(z)\mathrm{d}z=\int\lim_{t\to1-}h_{t}(z)\mathrm{d}z=\int\frac{\Vert z\Vert_{2}^{2}}{2}\phi(z)\mathrm{d}z=\frac{d}{2}.
\]
Therefore we have\begin{subequations}\label{eq:ub} 
\begin{equation}
\mathbb{E}[\log\rho_{t}(X_{t})\mymid X_{0}=x_{0}]\leq g_{x_{0}}(t)\qquad\text{where}\qquad g_{x_{0}}(t)\coloneqq\overline{g}_{x_{0}}(t)-\widetilde{g}_{x_{0}}(t),
\end{equation}
such that 
\begin{equation}
\lim_{t\to1-}g_{x_{0}}(t)=\lim_{t\to1-}\overline{g}_{x_{0}}(t)-\lim_{t\to1-}\widetilde{g}_{x_{0}}(t)=-\frac{d}{2}\log(2\pi)-\frac{d}{2}.
\end{equation}
\end{subequations}

\paragraph{Conclusion.}

By putting together (\ref{eq:lb}) and (\ref{eq:ub}), we know that
for any $t\in(0.9,1)$ 
\[
f_{x_{0}}(t)\leq\mathbb{E}[\log\rho_{t}(X_{t})\mymid X_{0}=x_{0}]\leq g_{x_{0}}(t)\qquad\text{and}\qquad\lim_{t\to1-}f_{x_{0}}(t)=\lim_{t\to1-}g_{x_{0}}(t)=-\frac{d}{2}\log(2\pi)-\frac{d}{2}.
\]
By the sandwich theorem, we arrive at the desired result 
\[
\lim_{t\to1-}\mathbb{E}[\log\rho_{t}(X_{t})\mymid X_{0}=x_{0}]=-\frac{d}{2}\log(2\pi)-\frac{d}{2}.
\]

\section{Proof of Proposition~\ref{prop:limit-0}\label{sec:proof-prop-limit-0}}

Suppose that $L\coloneqq\sup_{x}\Vert\nabla^{2}\log\rho_{0}(x)\Vert$.
The following claim will be useful in establishing the proposition,
whose proof is deferred to the end of this section.

\begin{claim}\label{claim:lispchitz} There exists some $t_{0}>0$
such that 
\begin{equation}
\sup_{x}\Vert\nabla^{2}\log\rho_{t}(x)\Vert\leq4L.\label{eq:proof-limit-0-1}
\end{equation}
holds for any $0\leq t\leq t_{0}$. \end{claim}

Equipped with Claim~\ref{claim:lispchitz}, we know that for any
$t\leq t_{0}$, 
\begin{align}
 & \mathbb{E}\big[\log\rho_{t}(X_{t})\mymid X_{0}=x_{0}\big]=\mathbb{E}\big[\log\rho_{t}(\sqrt{1-t}x_{0}+\sqrt{t}Z)\big]\nonumber \\
 & \qquad\overset{\text{(i)}}{=}\mathbb{E}\big[\log\rho_{t}(\sqrt{1-t}x_{0})+\sqrt{t}Z^{\top}\nabla\log\rho_{t}(\sqrt{1-t}x_{0})+O(Lt)\Vert Z\Vert_{2}^{2}\big]\nonumber \\
 & \qquad\overset{\text{(ii)}}{=}\log\rho_{t}(\sqrt{1-t}x_{0})+O(Ldt)\nonumber \\
 & \qquad\overset{\text{(iii)}}{=}\log\int_{x}\rho_{0}(x)(2\pi t)^{-d/2}\exp\Big(-\frac{(1-t)\|x-x_{0}\|_{2}^{2}}{2t}\Big)\mathrm{d}x+O(L\sqrt{dt})\nonumber \\
 & \qquad=(1-t)^{-d/2}\log\int_{x}\rho_{0}(x)\left(\frac{2\pi t}{1-t}\right)^{-d/2}\exp\Big(-\frac{(1-t)\|x-x_{0}\|_{2}^{2}}{2t}\Big)\mathrm{d}x+O(L\sqrt{dt}),\label{eq:proof-limit-0-2}
\end{align}
where $Z\sim\mathcal{N}(0,I_{d})$. Here step (i) follows from (\ref{eq:proof-limit-0-1})
in Claim~\ref{claim:lispchitz}; step (ii) holds since $\mathbb{E}[Z]=0$
and $\mathbb{E}[\Vert Z\Vert_{2}^{2}]=d$; while step (iii) follows
from (\ref{eq:pt-rescaled-density}). It is straightforward to check
that
\[
\int_{x}\rho_{0}(x)\left(\frac{2\pi t}{1-t}\right)^{-d/2}\exp\Big(-\frac{(1-t)\|x-x_{0}\|_{2}^{2}}{2t}\Big)\mathrm{d}x
\]
is the density of $\rho_{0}*\mathcal{N}(0,t/(1-t))$ evaluated at
$x_{0}$, which taken collectively with the assumption that $\rho_{0}(\cdot)$
is continuous yields
\[
\lim_{t\to0+}\int_{x}\rho_{0}(x)\left(\frac{2\pi t}{1-t}\right)^{-d/2}\exp\Big(-\frac{(1-t)\|x-x_{0}\|_{2}^{2}}{2t}\Big)\mathrm{d}x=\rho_{0}(x_{0}).
\]
Therefore we can take $t\to0+$ in (\ref{eq:proof-limit-0-2}) to
achieve
\[
\lim_{t\to0+}\mathbb{E}\big[\log\rho_{t}(X_{t})\mymid X_{0}=x_{0}\big]=\log\rho_{0}(x_{0})
\]
as claimed.

\paragraph*{Proof of Claim~\ref{claim:lispchitz}.}

The conditional density of $X_{0}$ given $X_{t}=x$ is
\begin{equation}
p_{X_{0}|X_{t}}(x_{0}\mymid x)=\frac{p_{X_{0}}(x_{0})p_{X_{t}|X_{0}}(x\mymid x_{0})}{p_{X_{t}}(x)}=\frac{\rho_{0}(x_{0})}{\rho_{t}(x)}(2\pi t)^{-d/2}\exp\left(-\frac{\Vert x-\sqrt{1-t}x_{0}\Vert_{2}^{2}}{2t}\right),\label{eq:proof-lipschitz-0}
\end{equation}
which leads to 
\begin{align*}
-\nabla_{x_{0}}^{2}\log p_{X_{0}|X_{t}}(x_{0}\mymid x) & =-\nabla_{x_{0}}^{2}\log\rho_{0}(x_{0})+\frac{1}{2t}\nabla_{x_{0}}^{2}\Vert x-\sqrt{1-t}x_{0}\Vert_{2}^{2}\\
 & =-\nabla_{x_{0}}^{2}\log\rho_{0}(x_{0})+\frac{1-t}{t}I_{d}\succeq\left(\frac{1-t}{t}-L\right)I_{d}.
\end{align*}
Therefore we know that
\begin{equation}
-\nabla_{x_{0}}^{2}\log p_{X_{0}|X_{t}}(x_{0}\mymid x)\succeq\frac{1}{2t}I_{d}\qquad\text{for}\qquad t\leq\frac{1}{2(L+1)},\label{eq:proof-lipschitz-1}
\end{equation}
namely the conditional distribution of $X_{0}$ given $X_{t}=x$ is
$1/(2t)$-strongly log-concave for any $x$, when $t\leq1/2(L+1)$.
By writting
\begin{align}
\rho_{t}(x) & =p_{X_{t}}(x)=\int\phi(z)p_{\sqrt{1-t}X_{0}}\left(x-\sqrt{t}z\right)\mathrm{d}z=(1-t)^{-d/2}\int\phi(z)\rho_{0}\left(\frac{x-\sqrt{t}z}{\sqrt{1-t}}\right)\mathrm{d}z,\label{eq:proof-lipschitz-2}
\end{align}
we can express the score function of $\rho_{t}$ as
\begin{align}
\nabla\log\rho_{t}(x) & =\frac{\nabla\rho_{t}(x)}{\rho_{t}(x)}=(1-t)^{-\frac{d+1}{2}}\frac{1}{\rho_{t}(x)}\int\phi(z)\nabla\rho_{0}\left(\frac{x-\sqrt{t}z}{\sqrt{1-t}}\right)\mathrm{d}z\nonumber \\
 & =(1-t)^{-\frac{d+1}{2}}\frac{1}{\rho_{t}(x)}\int\phi(z)\rho_{0}\left(\frac{x-\sqrt{t}z}{\sqrt{1-t}}\right)\nabla\log\rho_{0}\left(\frac{x-\sqrt{t}z}{\sqrt{1-t}}\right)\mathrm{d}z\label{eq:proof-lipschitz-3}\\
 & \overset{\text{(i)}}{=}(1-t)^{-\frac{d+1}{2}}\left(\frac{1-t}{t}\right)^{d/2}\frac{1}{\rho_{t}(x)}\int\phi\left(\frac{x-\sqrt{1-t}x_{0}}{\sqrt{t}}\right)\rho_{0}\left(x_{0}\right)\nabla\log\rho_{0}\left(x_{0}\right)\mathrm{d}x_{0}\nonumber \\
 & \overset{\text{(ii)}}{=}\frac{1}{\sqrt{1-t}}\int p_{X_{0}|X_{t}}(x_{0}\mymid x)\nabla\log\rho_{0}\left(x_{0}\right)\mathrm{d}x_{0}=\frac{1}{\sqrt{1-t}}\mathbb{E}\left[\nabla\log\rho_{0}\left(X_{0}\right)\mymid X_{t}=x\right].\label{eq:proof-lipschitz-4}
\end{align}
Here step (i) uses the change of variable $x_{0}=(x-\sqrt{t}z)/\sqrt{1-t}$,
while step (ii) follows from (\ref{eq:proof-lipschitz-0}). Starting
from (\ref{eq:proof-lipschitz-3}), we take the derivative to achieve
\begin{align}
\nabla^{2}\log\rho_{t}(x) & =\underbrace{(1-t)^{-\frac{d}{2}+1}\frac{1}{\rho_{t}(x)}\int\phi(z)\rho_{0}\left(\frac{x-\sqrt{t}z}{\sqrt{1-t}}\right)\nabla\log\rho_{0}\left(\frac{x-\sqrt{t}z}{\sqrt{1-t}}\right)\left[\nabla\log\rho_{0}\left(\frac{x-\sqrt{t}z}{\sqrt{1-t}}\right)\right]^{\top}\mathrm{d}z}_{\eqqcolon H_{1}(x)}\nonumber \\
 & \qquad+\underbrace{(1-t)^{-\frac{d}{2}+1}\frac{1}{\rho_{t}(x)}\int\phi(z)\rho_{0}\left(\frac{x-\sqrt{t}z}{\sqrt{1-t}}\right)\nabla^{2}\log\rho_{0}\left(\frac{x-\sqrt{t}z}{\sqrt{1-t}}\right)\mathrm{d}z}_{\eqqcolon H_{2}(x)}\nonumber \\
 & \qquad-\underbrace{(1-t)^{-\frac{d+1}{2}}\frac{1}{\rho_{t}^{2}(x)}\int\phi(z)\rho_{0}\left(\frac{x-\sqrt{t}z}{\sqrt{1-t}}\right)\nabla\log\rho_{0}\left(\frac{x-\sqrt{t}z}{\sqrt{1-t}}\right)\mathrm{d}z\left[\nabla\rho_{t}(x)\right]^{\top}}_{\eqqcolon H_{3}(x)}.\label{eq:proof-lipschitz-5}
\end{align}
Then we investigate $H_{1}(x)$, $H_{2}(x)$ and $H_{3}(x)$ respectively.
Regarding $H_{1}(x)$, we have\begin{subequations}\label{eq:proof-lipschitz-6}
\begin{align}
H_{1}(x) & \overset{\text{(a1)}}{=}(1-t)^{-\frac{d}{2}+1}\left(\frac{1-t}{t}\right)^{d/2}\frac{1}{\rho_{t}(x)}\int\phi\left(\frac{x-\sqrt{1-t}x_{0}}{\sqrt{t}}\right)\rho_{0}\left(x_{0}\right)\nabla\log\rho_{0}\left(x_{0}\right)\left[\nabla\log\rho_{0}\left(x_{0}\right)\right]^{\top}\mathrm{d}z\nonumber \\
 & \overset{\text{(b1)}}{=}\frac{1}{1-t}\int p_{X_{0}|X_{t}}(x_{0}\mymid x)\nabla\log\rho_{0}\left(x_{0}\right)\left[\nabla\log\rho_{0}\left(x_{0}\right)\right]^{\top}\mathrm{d}x_{0}\nonumber \\
 & =\frac{1}{1-t}\mathbb{E}\left[\nabla\log\rho_{0}\left(X_{0}\right)\left[\nabla\log\rho_{0}\left(X_{0}\right)\right]^{\top}\mymid X_{t}=x\right];
\end{align}
for $H_{2}(x)$, we have
\begin{align}
H_{2}(x) & \overset{\text{(a2)}}{=}(1-t)^{-\frac{d}{2}+1}\left(\frac{1-t}{t}\right)^{d/2}\frac{1}{\rho_{t}(x)}\int\phi\left(\frac{x-\sqrt{1-t}x_{0}}{\sqrt{t}}\right)\rho_{0}\left(x_{0}\right)\nabla^{2}\log\rho_{0}\left(\frac{x-\sqrt{t}z}{\sqrt{1-t}}\right)\mathrm{d}x_{0}\nonumber \\
 & \overset{\text{(b2)}}{=}\frac{1}{1-t}\int p_{X_{0}|X_{t}}(x_{0}\mymid x)\nabla^{2}\log\rho_{0}\left(x_{0}\right)\mathrm{d}x_{0}=\frac{1}{1-t}\mathbb{E}\left[\nabla^{2}\log\rho_{0}\left(X_{0}\right)\mymid X_{t}=x\right];
\end{align}
for the final term $H_{3}(x)$, we have
\begin{align}
H_{3}(x) & \overset{\text{(c)}}{=}-(1-t)^{-\frac{d+1}{2}}\frac{1}{\rho_{t}(x)}\left[\int\phi(z)\rho_{0}\left(\frac{x-\sqrt{t}z}{\sqrt{1-t}}\right)\nabla\log\rho_{0}\left(\frac{x-\sqrt{t}z}{\sqrt{1-t}}\right)\mathrm{d}z\right]\left[\nabla\log\rho_{t}(x)\right]^{\top}\nonumber \\
 & \overset{\text{(a3)}}{=}-(1-t)^{-\frac{d+1}{2}}\left(\frac{1-t}{t}\right)^{d/2}\frac{1}{\rho_{t}(x)}\left[\int\phi\left(\frac{x-\sqrt{1-t}x_{0}}{\sqrt{t}}\right)\rho_{0}\left(x_{0}\right)\nabla\log\rho_{0}\left(x_{0}\right)\mathrm{d}x_{0}\right]\left[\nabla\log\rho_{t}(x)\right]^{\top}\nonumber \\
 & \overset{\text{(b3)}}{=}-\frac{1}{\sqrt{1-t}}\int p_{X_{0}|X_{t}}(x_{0}\mymid x)\nabla\log\rho_{0}\left(x_{0}\right)\mathrm{d}x_{0}\left[\nabla\log\rho_{t}(x)\right]^{\top}\nonumber \\
 & \overset{\text{(d)}}{=}-\frac{1}{1-t}\mathbb{E}\left[\nabla\log\rho_{0}\left(X_{0}\right)\mymid X_{t}=x\right]\mathbb{E}\left[\nabla\log\rho_{0}\left(X_{0}\right)\mymid X_{t}=x\right]^{\top}.
\end{align}
\end{subequations}Here steps (a1), (a2) and (a3) follow from the
change of variable $x_{0}=(x-\sqrt{t}z)/\sqrt{1-t}$; steps (b1),
(b2) and (b3) utilize (\ref{eq:proof-lipschitz-0}); step (c) follows
from $\nabla\log\rho_{t}(x)=\nabla\rho_{t}(x)/\rho_{t}(x)$; while
step (d) follows from (\ref{eq:proof-lipschitz-4}). Substituting
(\ref{eq:proof-lipschitz-6}) back into (\ref{eq:proof-lipschitz-5}),
we have
\begin{equation}
\nabla^{2}\log\rho_{t}(x)=\frac{1}{1-t}\mathbb{E}\left[\nabla^{2}\log\rho_{0}\left(X_{0}\right)\mymid X_{t}=x\right]+\frac{1}{1-t}\mathsf{cov}\left(\nabla\log\rho_{0}\left(X_{0}\right)\mymid X_{t}=x\right).\label{eq:proof-lipschitz-7}
\end{equation}
Notice that for any $t\leq1/2(L+1)$, we have
\begin{align}
\left\Vert \mathsf{cov}\left(\nabla\log\rho_{0}\left(X_{0}\right)\mymid X_{t}=x\right)\right\Vert  & =\sup_{u\in\mathbb{S}^{d-1}}\mathbb{E}\left[\left[u^{\top}\left(\nabla\log\rho_{0}\left(X_{0}\right)-\mathbb{E}\left[\nabla\log\rho_{0}\left(X_{0}\right)\mymid X_{t}=x\right]\right)\right]^{2}\mymid X_{t}=x\right]\nonumber \\
 & \overset{\text{(i)}}{\leq}\sup_{u\in\mathbb{S}^{d-1}}\mathbb{E}\left[\left[u^{\top}\left(\nabla\log\rho_{0}\left(X_{0}\right)-\nabla\log\rho_{0}\left(\mathbb{E}\left[X_{0}\mymid X_{t}=x\right]\right)\right)\right]^{2}\mymid X_{t}=x\right]\nonumber \\
 & \leq\mathbb{E}\left[\left\Vert \nabla\log\rho_{0}\left(X_{0}\right)-\nabla\log\rho_{0}\left(\mathbb{E}\left[X_{0}\mymid X_{t}=x\right]\right)\right\Vert _{2}^{2}\mymid X_{t}=x\right]\nonumber \\
 & \overset{\text{(ii)}}{\leq}\mathbb{E}\left[\left\Vert X_{0}-\mathbb{E}\left[X_{0}\mymid X_{t}=x\right]\right\Vert _{2}^{2}\mymid X_{t}=x\right]\nonumber \\
 & \overset{\text{(iii)}}{\leq}2tL^{2}d,\label{eq:proof-lipschitz-8}
\end{align}
Here step (i) holds since for any random variable $X$, $\mathbb{E}[(X-c)^{2}]$
is minimized at $c=\mathbb{E}[X]$; step (ii) holds since the score
function $\nabla\log\rho_{0}(\cdot)$ is $L$-Lipschitz; step (iii)
follows from the Poincar\'e inequality for log-concave distribution,
and the fact that the conditional distribution of $X_{0}$ given $X_{t}=x$
is $1/2t$-strongly log-concave (cf.~(\ref{eq:proof-lipschitz-1})).
We conclude that
\begin{align*}
\left\Vert \nabla^{2}\log\rho_{t}(x)\right\Vert  & \overset{\text{(a)}}{\leq}\frac{1}{1-t}L+\frac{2tL^{2}d}{1-t}\overset{\text{(b)}}{\leq}4L.
\end{align*}
Here step (a) follows from (\ref{eq:proof-lipschitz-7}), (\ref{eq:proof-lipschitz-8}),
and the assumption that $\sup_{x}\Vert\nabla^{2}\log\rho_{t}(x)\Vert\leq L$,
while step (b) holds provided that $t\leq\min\{1/2,1/(2Ld)\}$.

\section{More discussions on the density formulas\label{appendix:discussion}}

Although the density formulas (\ref{eq:density-formula-smooth}) have
been rigorously established, it is helpful to inspect the limiting
behavior of the integrand $D(t,x_{0})$ at the boundary to understand
why the integral converges. Throughout the discussion, we let $\varepsilon\sim\mathcal{N}(0,I_{d})$. 
\begin{itemize}
\item As $t\to0$, we can compute 
\begin{align*}
D(t,x_{0}) & \asymp\frac{\mathbb{E}\big[\Vert\varepsilon+\sqrt{t}\nabla\log\rho_{t}(\sqrt{1-t}x_{0}+\sqrt{t}\varepsilon)\Vert_{2}^{2}\big]-d}{t}\\
 & \overset{\text{(i)}}{\asymp}\mathbb{E}\big[\Vert\nabla\log\rho_{t}(\sqrt{1-t}x_{0}+\sqrt{t}\varepsilon)\Vert_{2}^{2}\big]+\frac{1}{\sqrt{t}}\mathbb{E}\big[\varepsilon^{\top}\nabla\log\rho_{t}(\sqrt{1-t}x_{0}+\sqrt{t}\varepsilon)\big]\\
 & \overset{\text{(ii)}}{\asymp}\mathbb{E}\big[\Vert\nabla\log\rho_{t}(\sqrt{1-t}x_{0}+\sqrt{t}\varepsilon)\Vert_{2}^{2}\big]+\mathbb{E}\left[\mathsf{tr}\left(\nabla^{2}\log\rho_{t}(\sqrt{1-t}x_{0}+\sqrt{t}\varepsilon)\right)\right].
\end{align*}
Here step (i) holds since $\mathbb{E}[\Vert\varepsilon\Vert_{2}^{2}]=d$,
while step (ii) follows from Stein's lemma. Therefore, when the score
functions are reasonably smooth as $t\to0$, one may expect that the
integrand $D(t,x_{0})$ is of constant order, allowing the integral
to converge at $t=0$. 
\item As $t\to1$, we can compute 
\begin{align*}
D(t,x_{0}) & =\frac{1}{2(1-t)t}\mathbb{E}\big[\Vert\varepsilon+\sqrt{t}\nabla\log\rho_{t}(\sqrt{1-t}x_{0}+\sqrt{t}\varepsilon)\Vert_{2}^{2}\big]-\frac{d}{2t}\\
 & \asymp\frac{1}{2(1-t)}\mathbb{E}\big[\Vert\varepsilon+\sqrt{t}\nabla\log\rho_{t}(\sqrt{1-t}x_{0}+\sqrt{t}\varepsilon)\Vert_{2}^{2}\big]-\frac{d}{2}.
\end{align*}
Since $\rho_{t}$ converges to $\phi$ as $t\to1$ and $\nabla\log\phi(x)=-x$,
we have 
\[
\lim_{t\to1}\varepsilon+\sqrt{t}\nabla\log\rho_{t}(\sqrt{1-t}x_{0}+\sqrt{t}\varepsilon)=0.
\]
Hence one may expect that $\mathbb{E}\big[\Vert\varepsilon+\sqrt{t}\nabla\log\rho_{t}(\sqrt{1-t}x_{0}+\sqrt{t}\varepsilon)\Vert_{2}^{2}\big]$
converges to zero quickly, allowing the integral to converge at $t=1$. 
\end{itemize}

\section{Technical details in Section~\ref{sec:implications} \label{appendix:technical}}

\subsection{Technical details in Section~\ref{subsec:vlb} \label{appendix:vlb}}

\paragraph{Computing $L_{t-1}(x_{0})$.}

Conditional on $X_{t}=x_{t}$ and $X_{0}=x_{0}$, we have 
\[
X_{t-1}\,|\,X_{t}=x_{t},X_{0}=x_{0}\sim\mathcal{N}\left(\frac{\sqrt{\overline{\alpha}_{t-1}}\beta_{t}}{1-\overline{\alpha}_{t}}x_{0}+\frac{\sqrt{\alpha_{t}}(1-\overline{\alpha}_{t-1})}{1-\overline{\alpha}_{t}}x_{t},\frac{1-\overline{\alpha}_{t-1}}{1-\overline{\alpha}_{t}}\beta_{t}I_{d}\right),
\]
and conditional on $Y_{t}=x_{t}$, we have 
\[
Y_{t-1}\,|\,Y_{t}=x_{t}\sim\mathcal{N}\left(\frac{x_{t}+\eta_{t}s_{t}\left(x_{t}\right)}{\sqrt{\alpha_{t}}},\frac{\sigma_{t}^{2}}{\alpha_{t}}\right).
\]
Recall that the KL divergence between two $d$-dimensional Gaussian
$\mathcal{N}(\mu_{1},\Sigma_{1})$ and $\mathcal{N}(\mu_{2},\Sigma_{2})$
admits the following closed-form expression: 
\[
\mathsf{KL}\left(\mathcal{N}(\mu_{1},\Sigma_{1})\,\Vert\,\mathcal{N}(\mu_{2},\Sigma_{2})\right)=\frac{1}{2}\left[\mathsf{tr}\left(\Sigma_{2}^{-1}\Sigma_{1}\right)+\left(\mu_{2}-\mu_{1}\right)^{\top}\Sigma_{2}^{-1}\left(\mu_{2}-\mu_{1}\right)-d+\log\det\Sigma_{2}-\log\det\Sigma_{1}\right].
\]
Then we can check that for $2\leq t\leq T$, 
\[
\mathsf{KL}\big(p_{X_{t-1}|X_{t},X_{0}}(\cdot\mymid x_{t},x_{0})\,\Vert\,p_{Y_{t-1}|Y_{t}}(\cdot\mymid x_{t})\big)=\frac{\alpha_{t}}{2\sigma_{t}^{2}}\left\Vert \frac{\sqrt{\overline{\alpha}_{t-1}}\beta_{t}}{1-\overline{\alpha}_{t}}x_{0}+\frac{\alpha_{t}-1}{\sqrt{\alpha_{t}}(1-\overline{\alpha}_{t})}x_{t}-\frac{\eta_{t}s_{t}(x_{t})}{\sqrt{\alpha_{t}}}\right\Vert _{2}^{2},
\]
where we use the coefficient design (\ref{eq:defn-step-size}). This
immediately gives 
\begin{align*}
L_{t-1}(x_{0}) & =\frac{\alpha_{t}}{2\sigma_{t}^{2}}\mathbb{E}_{x_{t}\sim p_{X_{t}|X_{0}}(\cdot\mymid x_{0})}\left[\left\Vert \frac{\sqrt{\overline{\alpha}_{t-1}}\beta_{t}}{1-\overline{\alpha}_{t}}x_{0}+\frac{\alpha_{t}-1}{\sqrt{\alpha_{t}}(1-\overline{\alpha}_{t})}x_{t}-\frac{\eta_{t}s_{t}(x_{t})}{\sqrt{\alpha_{t}}}\right\Vert _{2}^{2}\right]\\
 & \overset{\text{(i)}}{=}\frac{\alpha_{t}}{2\sigma_{t}^{2}}\mathbb{E}_{\varepsilon\sim\mathcal{N}(0,I_{d})}\left[\bigg\Vert\frac{\alpha_{t}-1}{\sqrt{\alpha_{t}(1-\overline{\alpha}_{t})}}\varepsilon-\frac{1-\alpha_{t}}{\sqrt{\alpha_{t}}}s_{t}(\sqrt{\overline{\alpha}_{t}}x_{0}+\sqrt{1-\overline{\alpha}_{t}}\varepsilon)\bigg\Vert_{2}^{2}\right]\\
 & \overset{\text{(ii)}}{=}\frac{1-\alpha_{t}}{2(\alpha_{t}-\overline{\alpha}_{t})}\mathbb{E}_{\varepsilon\sim\mathcal{N}(0,I_{d})}\left[\left\Vert \varepsilon-\varepsilon_{t}(\sqrt{\overline{\alpha}_{t}}x_{0}+\sqrt{1-\overline{\alpha}_{t}}\varepsilon)\right\Vert _{2}^{2}\right].
\end{align*}
Here in step (i), we utilize the coefficient design (\ref{eq:defn-step-size})
and replace $x_{t}$ with $\sqrt{\overline{\alpha}_{t}}x_{0}+\sqrt{1-\overline{\alpha}_{t}}\varepsilon$,
which has the same distribution; while in step (ii), we replace the
score function $s_{t}(\cdot)$ with the epsilon predictor $\varepsilon_{t}(\cdot)\coloneqq-\sqrt{1-\overline{\alpha}_{t}}s_{t}(\cdot)$.
Comparing the coefficients in $L_{t-1}^{\star}$ and $L_{t-1}$, we
decompose 
\[
\left|\frac{1-\alpha_{t+1}}{2(1-\overline{\alpha}_{t})}-\frac{1-\alpha_{t}}{2(\alpha_{t}-\overline{\alpha}_{t})}\right|\leq\underbrace{\left|\frac{1-\alpha_{t+1}}{2(1-\overline{\alpha}_{t})}-\frac{1-\alpha_{t+1}}{2(\alpha_{t}-\overline{\alpha}_{t})}\right|}_{\eqqcolon\gamma_{1}}+\underbrace{\left|\frac{1-\alpha_{t+1}}{2(\alpha_{t}-\overline{\alpha}_{t})}-\frac{1-\alpha_{t}}{2(\alpha_{t}-\overline{\alpha}_{t})}\right|}_{\eqqcolon\gamma_{2}}.
\]
Consider the learning rate schedule in \citet{li2023towards,li2024adapting}:
\begin{equation}
\beta_{1}=\frac{1}{T^{c_{0}}},\qquad\beta_{t+1}=\frac{c_{1}\log T}{T}\min\left\{ \beta_{1}\left(1+\frac{c_{1}\log T}{T}\right)^{t},1\right\} \quad(t=1,\ldots,T-1)\label{eq:learning-rates}
\end{equation}
for sufficiently large constants $c_{0},c_{1}>0$. Then using the
properties in e.g., \citet[Lemma 8]{li2024adapting}, we can check
that 
\begin{align*}
\gamma_{1} & =\left|\frac{(1-\alpha_{t+1})(\alpha_{t}-1)}{2(1-\overline{\alpha}_{t})(\alpha_{t}-\overline{\alpha}_{t})}\right|\leq\frac{8c_{1}\log T}{T}\left|\frac{1-\alpha_{t+1}}{2(1-\overline{\alpha}_{t})}\right|,
\end{align*}
and 
\begin{align*}
\gamma_{2} & =\left|\frac{\alpha_{t}-\alpha_{t+1}}{2(\alpha_{t}-\overline{\alpha}_{t})}\right|=\left|\frac{\beta_{t}-\beta_{t+1}}{2(\alpha_{t}-\overline{\alpha}_{t})}\right|\leq\left|1-\frac{\beta_{t}}{\beta_{t+1}}\right|\left|1+\frac{1-\alpha_{t}}{\alpha_{t}-\overline{\alpha}_{t}}\right|\left|\frac{1-\alpha_{t+1}}{2(1-\overline{\alpha}_{t})}\right|\leq\frac{8c_{1}\log T}{T}\left|\frac{1-\alpha_{t+1}}{2(1-\overline{\alpha}_{t})}\right|.
\end{align*}
Hence the coefficients in $L_{t-1}^{\star}$ and $L_{t-1}$ are identical
up to higher-order error: 
\[
\left|\frac{1-\alpha_{t+1}}{2(1-\overline{\alpha}_{t})}-\frac{1-\alpha_{t}}{2(\alpha_{t}-\overline{\alpha}_{t})}\right|\leq\frac{16c_{1}\log T}{T}\left|\frac{1-\alpha_{t+1}}{2(1-\overline{\alpha}_{t})}\right|.
\]

\paragraph{Computing $L_{0}(x_{0})$.}

By taking $\eta_{1}=\sigma_{1}^{2}=1-\alpha_{1}$ (notice that (\ref{eq:defn-step-size})
does not cover the case $t=1$), we have 
\begin{align*}
p_{Y_{0}|Y_{1}}(x_{0}\mymid x_{1}) & =\left(\frac{2\pi\sigma_{1}^{2}}{\alpha_{1}}\right)^{-d/2}\exp\left(-\frac{\alpha_{1}}{2\sigma_{1}^{2}}\left\Vert x_{0}-\frac{x_{1}-\eta_{1}s_{1}\left(x_{1}\right)}{\sqrt{\alpha_{1}}}\right\Vert _{2}^{2}\right)\\
 & =\left(\frac{2\pi\beta_{1}}{\alpha_{1}}\right)^{-d/2}\exp\left(-\frac{\alpha_{1}}{2\beta_{1}}\left\Vert x_{0}-\frac{x_{1}-\beta_{1}s_{1}\left(x_{1}\right)}{\sqrt{\alpha_{1}}}\right\Vert _{2}^{2}\right),
\end{align*}
and therefore 
\begin{align}
C_{0}(x_{0}) & =\mathbb{E}_{x_{1}\sim p_{X_{1}|X_{0}}(\cdot\mymid x_{0})}\left[-\frac{d}{2}\log\frac{2\pi\beta_{1}}{\alpha_{1}}-\frac{\alpha_{1}}{2\beta_{1}}\left\Vert x_{0}-\frac{x_{1}+\beta_{1}s_{1}(x_{1})}{\sqrt{\alpha_{1}}}\right\Vert _{2}^{2}\right]\nonumber \\
 & \overset{\text{(i)}}{=}-\frac{d}{2}\log\frac{2\pi\beta_{1}}{\alpha_{1}}-\frac{1}{2}\mathbb{E}_{\varepsilon\sim\mathcal{N}(0,I_{d})}\left[\Vert\varepsilon+\sqrt{\beta_{1}}s_{1}(\sqrt{1-\beta_{1}}x_{0}+\sqrt{\beta_{1}}\varepsilon)\Vert_{2}^{2}\right]\nonumber \\
 & \overset{\text{(ii)}}{=}-\frac{1+\log(2\pi\beta_{1})}{2}d+\frac{d}{2}\log(1-\beta_{1})-\frac{1}{2}\beta_{1}\mathbb{E}_{\varepsilon\sim\mathcal{N}(0,I_{d})}\big[\Vert s_{1}(\sqrt{1-\beta_{1}}x_{0}+\sqrt{\beta_{1}}\varepsilon)\Vert_{2}^{2}\big]\nonumber \\
 & \quad\qquad-\sqrt{\beta_{1}}\mathbb{E}_{\varepsilon\sim\mathcal{N}(0,I_{d})}\big[\varepsilon^{\top}s_{1}(\sqrt{1-\beta_{1}}x_{0}+\sqrt{\beta_{1}}\varepsilon)\big].\label{eq:L0-1}
\end{align}
Here in step (i), we replace $x_{1}$ with $\sqrt{1-\beta_{1}}x_{0}+\sqrt{\beta_{1}}\varepsilon$,
which has the same distribution; step (ii) uses the fact that $\mathbb{E}[\Vert\varepsilon\Vert_{2}^{2}]=d$
for $\varepsilon\sim\mathcal{N}(0,I_{d})$. Using similar analysis
as in Proposition~\ref{prop:limit-0}, we can show that $\sup_{x}\Vert\nabla^{2}\log q_{1}(x)\Vert\leq O(L)$
when $\beta_{1}$ is sufficiently small, as long as $\sup_{x}\Vert\nabla^{2}\log q_{0}(x)\Vert\leq L$.
Hence we have 
\begin{align}
\mathbb{E}_{\varepsilon\sim\mathcal{N}(0,I_{d})}\big[\Vert s_{1}(\sqrt{1-\beta_{1}}x_{0}+\sqrt{\beta_{1}}\varepsilon)\Vert_{2}^{2}\big] & \leq\mathbb{E}_{\varepsilon\sim\mathcal{N}(0,I_{d})}\big[\big(\Vert s_{1}(x_{0})\Vert_{2}+O(L)\Vert x_{0}-\sqrt{1-\beta_{1}}x_{0}-\sqrt{\beta_{1}}\varepsilon\Vert_{2}\big)^{2}\big]\nonumber \\
 & \leq2\Vert s_{1}(x_{0})\Vert_{2}^{2}+O(L^{2})\mathbb{E}_{\varepsilon\sim\mathcal{N}(0,I_{d})}\big[\Vert x_{0}-\sqrt{1-\beta_{1}}x_{0}-\sqrt{\beta_{1}}\varepsilon\Vert_{2}^{2}\big]\nonumber \\
 & =2\left\Vert s_{1}(x_{0})\right\Vert _{2}^{2}+O(L^{2}\beta_{1}).\label{eq:L0-2}
\end{align}
By Stein's lemma, we can show that 
\begin{align}
\mathbb{E}_{\varepsilon\sim\mathcal{N}(0,I_{d})}\left[\varepsilon^{\top}s_{1}(\sqrt{1-\beta_{1}}x_{0}+\sqrt{\beta_{1}}\varepsilon)\right] & =\sqrt{\beta_{1}}\mathbb{E}\left[\mathsf{tr}\left(\nabla^{2}\log q_{1}(\sqrt{1-\beta_{1}}x_{0}+\sqrt{\beta_{1}}\varepsilon)\right)\right]\nonumber \\
 & \leq O(\sqrt{\beta_{1}}Ld).\label{eq:L0-3}
\end{align}
Substituting the bounds (\ref{eq:L0-2}) and (\ref{eq:L0-3}) back
into (\ref{eq:L0-1}), we have 
\[
C_{0}(x_{0})=-\frac{1+\log(2\pi\beta_{1})}{2}d+O(\beta_{1})
\]
as claimed.

\paragraph{Negligibility of $L_{T}(x)$. }

Since
\[
Y_{T}\sim\mathcal{N}(0,I_{d}),\qquad\text{and}\qquad X_{T}\mymid X_{0}=x_{0}\sim\mathcal{N}\left(\sqrt{\overline{\alpha}_{T}}x_{0},(1-\overline{\alpha}_{T})I_{d}\right),
\]
we can compute
\[
\mathsf{KL}\left(p_{Y_{T}}(\cdot)\,\Vert\,p_{X_{T}|X_{0}}(\cdot\mymid x_{0})\right)=\frac{1}{2}\frac{\overline{\alpha}_{T}}{1-\overline{\alpha}_{T}}\left(d+\Vert x_{0}\Vert_{2}^{2}\right)+\frac{d}{2}\log(1-\overline{\alpha}_{T})\leq\frac{1}{2}\frac{\overline{\alpha}_{T}}{1-\overline{\alpha}_{T}}\left(d+\Vert x_{0}\Vert_{2}^{2}\right).
\]
Using the learning rate schedule in \eqref{eq:learning-rates}, we can check
that $\overline{\alpha}_{T}\leq T^{-c_{2}}$ for some large universal
constant $c_{2}>0$; see e.g., \citet[Section 5.1]{li2023towards} for the proof.
Therefore when $T\geq2$, we have
\[
\mathsf{KL}\left(p_{Y_{T}}(\cdot)\,\Vert\,p_{X_{T}|X_{0}}(\cdot\mymid x_{0})\right)\leq\frac{d+\Vert x_{0}\Vert_{2}^{2}}{4T^{c_{2}}},
\]
which is negligible when $T$ is sufficiently large.

\paragraph{Optimal solution for \eqref{eq:expected-elbo}.}

It is known that for each $1\leq t\leq T$, the score function $s_{t}^{\star}(\cdot)$
associated with $q_{t}$ satisfies
\[
s_{t}^{\star}(\cdot)=\mathop{\arg\min}_{s(\cdot):\mathbb{R}^{d}\to\mathbb{R}^{d}}\mathbb{E}_{x\sim q_{0},\varepsilon\sim\mathcal{N}(0,I_{d})}\left[\left\Vert s\left(\sqrt{\overline{\alpha}_{t}}x+\sqrt{1-\overline{\alpha}_{t}}\varepsilon\right)+\frac{1}{\sqrt{1-\overline{\alpha}_{t}}}\varepsilon\right\Vert _{2}^{2}\right].
\]
See e.g., \citet[Appendix A]{chen2022sampling} for the proof. Recall that $\varepsilon_{t}^{\star}(\cdot)=\sqrt{1-\overline{\alpha}_{t}}s_{t}^{\star}(\cdot)$,
then we have
\[
\varepsilon_{t}^{\star}(\cdot)=\mathop{\arg\min}_{\varepsilon(\cdot):\mathbb{R}^{d}\to\mathbb{R}^{d}}\mathbb{E}_{x\sim q_{0},\varepsilon\sim\mathcal{N}(0,I_{d})}\Big[\big\Vert\varepsilon-\varepsilon(\sqrt{\overline{\alpha}_{t}}x+\sqrt{1-\overline{\alpha}_{t}}\varepsilon)\big\Vert_{2}^{2}\Big].
\]
Therefore the global minimizer for \eqref{eq:expected-elbo} is
$\widehat{\varepsilon}_{t}(\cdot)\equiv\varepsilon_{t}^{\star}(\cdot)$
for each $1\leq t\leq T$.

\subsection{Technical details in Section~\ref{subsec:gan} \label{appendix:gan}}

By checking the optimality condition, we know that $(D_{\lambda},G_{\lambda})$
is a Nash equilibrium if and only if 
\begin{align}
D_{\lambda}(x) & =\frac{p_{\mathsf{data}}(x)}{p_{\mathsf{data}}(x)+p_{G_{\lambda}}(x)},\qquad\text{(optimality condition for }D_{\lambda})\label{eq:opt-D}
\end{align}
where $p_{G_{\lambda}}=(G_{\lambda})_{\#}p_{\mathsf{noise}}$, and
there exists some constant $c$ such that 
\begin{equation}
\begin{cases}
-\log D_{\lambda}(x)+\lambda L(x)=c, & \text{when}\quad x\in\mathsf{supp}(p_{G_{\lambda}}),\\
-\log D_{\lambda}(x)+\lambda L(x)\geq c, & \text{otherwise}.
\end{cases}\qquad(\text{optimality condition for }G_{\lambda})\label{eq:opt-G}
\end{equation}
Taking the approximation $L(x)\approx-\log p_{\mathsf{data}}(x)+C_{0}^{\star}$
as exact, we have 
\begin{equation}
D_{\lambda}(x)=\begin{cases}
e^{\lambda C_{0}^{\star}-c}p_{\mathsf{data}}^{-\lambda}(x), & \text{for }x\in\mathsf{supp}(p_{G_{\lambda}}),\\
1, & \text{for }x\notin\mathsf{supp}(p_{G_{\lambda}}).
\end{cases}\label{eq:D-equality}
\end{equation}
where the first and second cases follow from (\ref{eq:opt-G}) and
(\ref{eq:opt-D}) respectively. Then we derive a closed-form expression
for $p_{G_{\lambda}}$. 
\begin{itemize}
\item For any $x\in\mathsf{supp}(p_{G_{\lambda}})$, by putting (\ref{eq:opt-D})
and (\ref{eq:D-equality}) together, we have 
\[
e^{\lambda C_{0}^{\star}-c}p_{\mathsf{data}}^{-\lambda}(x)=\frac{p_{\mathsf{data}}(x)}{p_{\mathsf{data}}(x)+p_{G_{\lambda}}(x)},
\]
which further gives 
\begin{equation}
p_{G_{\lambda}}(x)=p_{\mathsf{data}}(x)\big(e^{-\lambda C_{0}^{\star}+c}p_{\mathsf{data}}^{\lambda}(x)-1\big).\label{eq:p-G-case-1}
\end{equation}
\item For any $x\notin\mathsf{supp}(p_{G_{\lambda}})$, we have 
\[
-\log D_{\lambda}(x)+\lambda L(x)\overset{\text{(i)}}{=}\lambda L(x)\overset{\text{(ii)}}{=}-\lambda\log p_{\mathsf{data}}(x)+\lambda C_{0}^{\star}\overset{\text{(iii)}}{\geq}c,
\]
where step (i) follows from $D_{\lambda}(x)=1$, which follows from
(\ref{eq:D-equality}); step (ii) holds when we take the approximation
$L(x)\approx-\log p_{\mathsf{data}}(x)+C_{0}^{\star}$ as exact; and
step (iii) follows from (\ref{eq:opt-G}). This immediately gives
\begin{equation}
e^{-\lambda C_{0}^{\star}+c}p_{\mathsf{data}}^{\lambda}(x)-1=\log\left(-\lambda C_{0}^{\star}+c+\lambda\log p_{\mathsf{data}}(x)\right)-1\leq0.\label{eq:p-G-case-2}
\end{equation}
\end{itemize}
Taking (\ref{eq:p-G-case-1}) and (\ref{eq:p-G-case-2}) collectively,
we can write 
\begin{equation}
p_{G_{\lambda}}(x)=p_{\mathsf{data}}(x)\big(e^{-\lambda C_{0}^{\star}+c}p_{\mathsf{data}}^{\lambda}(x)-1\big)_{+}.\label{eq:p-G-equality}
\end{equation}
On the other hand, we can check that (\ref{eq:D-equality}) and (\ref{eq:p-G-equality})
satisfies the optimality conditions (\ref{eq:opt-D}) and (\ref{eq:opt-G}),
which establishes the desired result. 
\bibliographystyle{apalike}
\bibliography{reference-diffusion}

\end{document}